\documentclass[10pt,twocolumn,letterpaper]{article}

\usepackage[pagenumbers]{cvpr} 

\usepackage[table, dvipsnames, pdftex]{xcolor}  
\usepackage{blindtext} 
\usepackage{float}
\usepackage{booktabs}
\usepackage{xargs}
\usepackage{pgf}
\usepackage{microtype}
\usepackage{multirow}
\newcommandx{\unsure}[2][1=]{\todo[linecolor=red,backgroundcolor=red!25,bordercolor=red,#1]{#2}}
\newcommandx{\change}[2][1=]{\todo[linecolor=blue,backgroundcolor=blue!25,bordercolor=blue,#1]{#2}}
\newcommandx{\info}[2][1=]{\todo[linecolor=OliveGreen,backgroundcolor=OliveGreen!25,bordercolor=OliveGreen,#1]{#2}}
\newcommandx{\improvement}[2][1=]{\todo[linecolor=Plum,backgroundcolor=Plum!25,bordercolor=Plum,#1]{#2}}
\newcommandx{\thiswillnotshow}[2][1=]{\todo[disable,#1]{#2}}
\definecolor{tabfirst}{rgb}{1, 0.7, 0.7} 
\definecolor{tabsecond}{rgb}{1, 0.85, 0.7} 
\definecolor{tabthird}{rgb}{1, 1, 0.7} 
\definecolor{tabrealtime}{rgb}{0.8, 1, 0.8} 
\definecolor{surfred}{rgb}{0.72, 0.33, 0.31}
\definecolor{surfgreen}{rgb}{0.51, 0.70, 0.4}
\definecolor{surfblue}{rgb}{0.423, 0.5568, 0.7490}

\definecolor{cvprblue}{rgb}{0.21,0.49,0.74}
\usepackage[pagebackref,breaklinks,colorlinks,allcolors=cvprblue]{hyperref}

\newcommand{\boldparagraph}[1]{\vspace{0.2cm}\noindent{\bf #1:} }

\newcommand{\beginsupplement}{%
    \setcounter{table}{0}
    \renewcommand{\thetable}{S\arabic{table}}%
    \setcounter{figure}{0}
    \renewcommand{\thefigure}{S\arabic{figure}}%
    \setcounter{section}{0}
    \renewcommand{\thesection}{S\arabic{section}}%
    \setcounter{equation}{0}%
    \renewcommand{\theequation}{S\arabic{equation}}%
}

\DeclareCaptionLabelFormat{mysubfig}{$#2$) }

\def\paperID{12831} 
\def\confName{CVPR\xspace}
\def\confYear{2025\xspace}

\title{Volumetric Surfaces: Representing Fuzzy Geometries with Layered Meshes}

\author{
   Stefano Esposito$^1$ \quad
   Anpei Chen$^1$ \quad
   Christian Reiser$^1$ \quad
   Samuel Rota Bulò$^2$ \quad
   Lorenzo Porzi$^2$ \\[0.25em]
   Katja Schwarz$^2$ \quad
   Christian Richardt$^2$ \quad
   Michael Zollh\"ofer$^2$ \quad
   Peter Kontschieder$^2$ \quad 
   Andreas Geiger$^1$ \\[0.75em]
   $^1$University of T\"ubingen \quad 
   $^2$Meta Reality Labs \\
}

\begin{document}


%
%
%
%
%

\twocolumn[{
\renewcommand\twocolumn[1][]{#1}%
\maketitle
\centering
\begin{minipage}{0.33\linewidth}
  \begin{center}
    \includegraphics[width=0.85\linewidth]{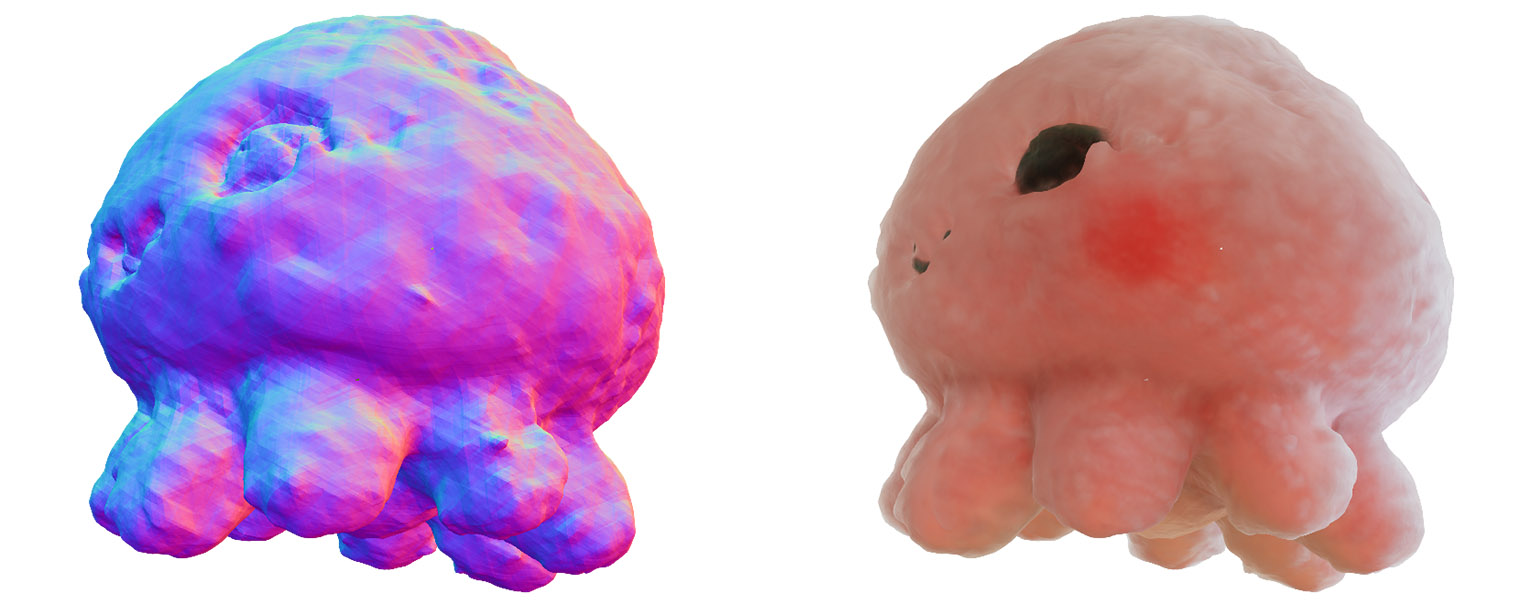} \\ 
    \emph{a}) PermutoSDF \vspace{3pt}
  \end{center}
  \vspace{-20pt}
  \begin{center}
    \includegraphics[width=0.85\linewidth]{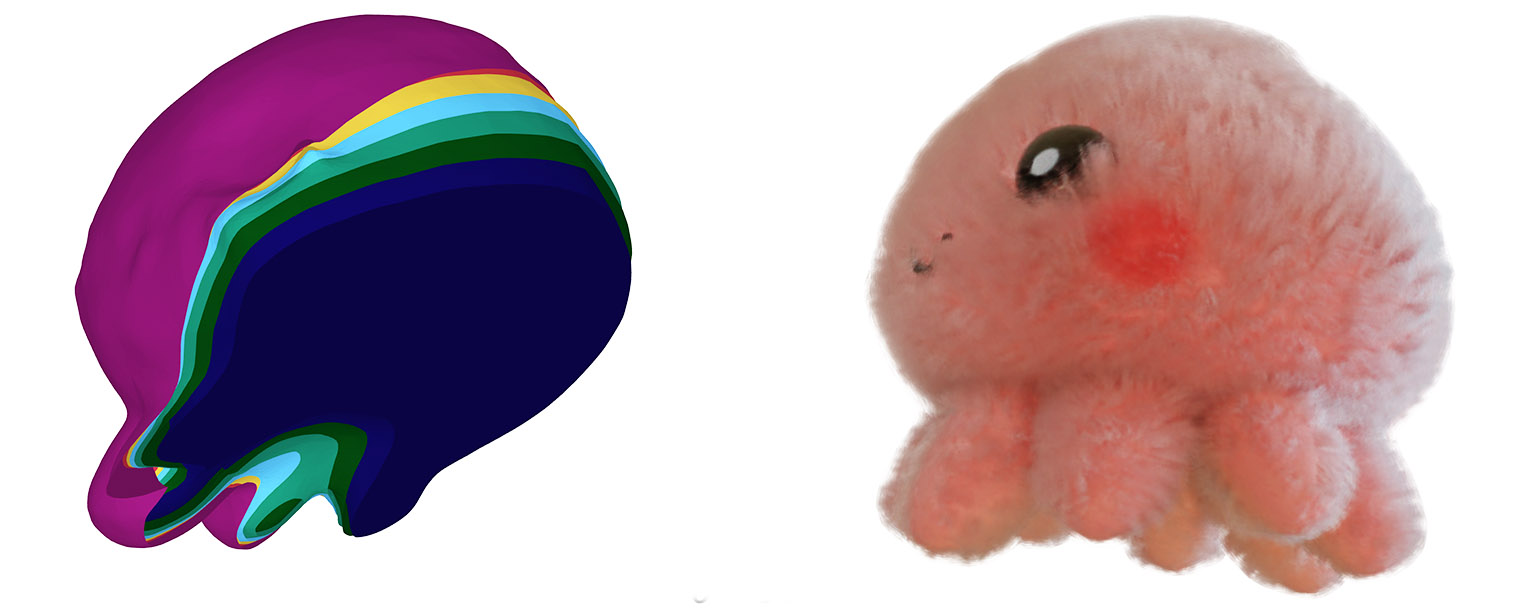} \\ 
    \emph{b}) Ours (7-Mesh) \vspace{3pt}
  \end{center}
\end{minipage}
\begin{minipage}{0.33\linewidth}
  \begin{center}
    \includegraphics[width=0.745\linewidth]{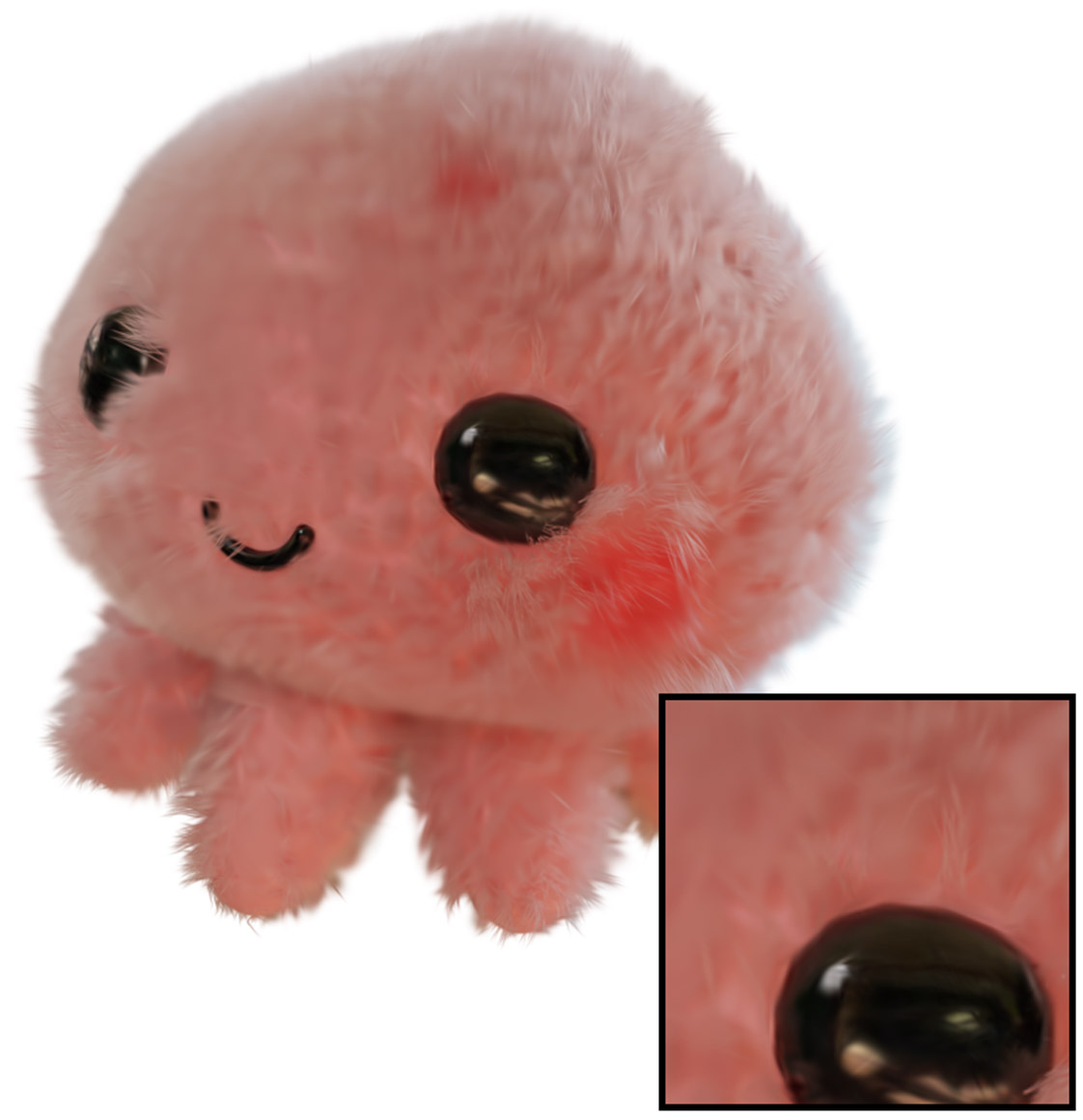} \\
    \emph{c}) 3DGS
  \end{center}
\end{minipage}
\begin{minipage}{0.33\linewidth}
\begin{center}
    \includegraphics[width=0.745\linewidth]{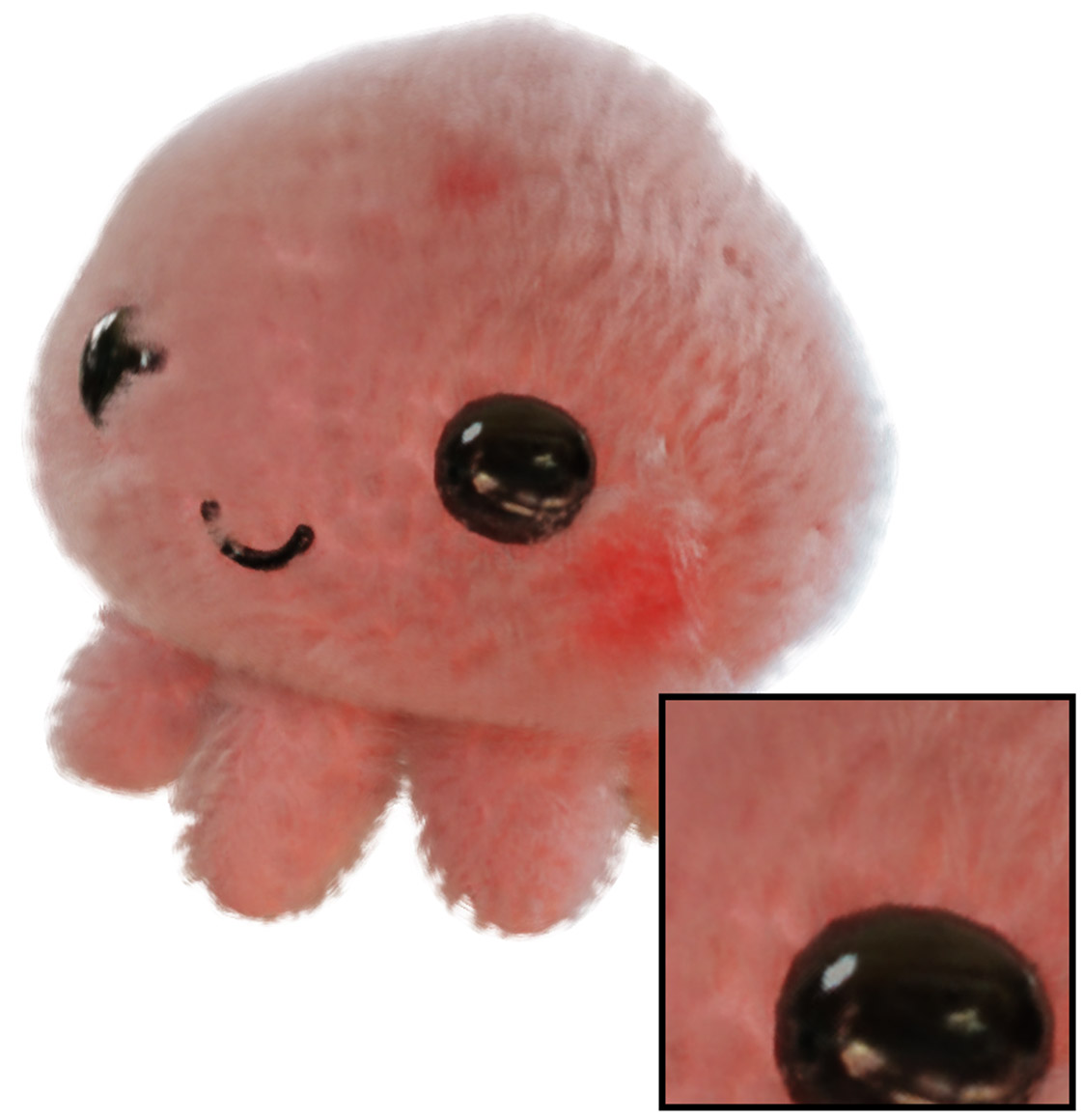} \\
    \emph{d}) Ours (7-Mesh)
\end{center}
\end{minipage}
\captionsetup{type=figure}
\caption{
    Our Volumetric Surfaces representation (\emph{b}) consists of $k$ lightweight, semi-transparent mesh shells that efficiently render fuzzy geometries with a limited number of samples $(3 \!\leq\! k \!\leq\! 9)$ via rasterization.
    Our image quality surpasses that of surface-based methods (\emph{a}) and approaches the quality of 3D Gaussian splatting (\emph{c}).
    Unlike splatting methods, our sorting-free representation enables faster rendering, particularly on low-power or mobile graphics hardware. 
    Project page: \href{https://autonomousvision.github.io/volsurfs}{https://autonomousvision.github.io/volsurfs}.
    \vspace{2em}
}
\label{fig:teaser}
}]


\begin{abstract}
High-quality view synthesis relies on volume rendering, splatting, or surface rendering. While surface rendering is typically the fastest, it struggles to accurately model fuzzy geometry like hair.
In turn, alpha-blending techniques excel at representing fuzzy materials but require an unbounded number of samples per ray (P1).
Further overheads are induced by empty space skipping in volume rendering (P2) and sorting input primitives in splatting (P3).
We present a novel representation for real-time view synthesis where the (P1) number of sampling locations is small and bounded, (P2) sampling locations are efficiently found via rasterization, and (P3) rendering is sorting-free.
We achieve this by representing objects as semi-transparent multi-layer meshes rendered in a fixed order.
First, we model surface layers as signed distance function (SDF) shells with optimal spacing learned during training.
Then, we bake them as meshes and fit UV textures. 
Unlike single-surface methods, our multi-layer representation effectively models fuzzy objects. 
In contrast to volume and splatting-based methods, our approach enables real-time rendering on low-power laptops and smartphones.
\end{abstract}    
\section{Introduction}
\label{sec:introduction}

Real-time rendering on mobile devices is challenging due to limited processing power, memory, and thermal constraints.
Recent methods for real-time view synthesis can be categorized according to their rendering paradigm.
On the one hand, we have surface-based methods like BakedSDF \cite{Yariv2023ARXIV} or BOG \cite{Reiser2024SIGGRAPH}, where rendering a pixel requires only reading appearance data from a single sampling location along the ray.
On the other hand, we have volume-based methods like SMERF \cite{Duckworth2023TOG} or 3DGS \cite{Kerbl2023SIGGRAPH}, where rendering a pixel requires reading data from multiple sample locations along the ray.
As a result, surface-based methods are generally faster than volume-based techniques, which struggle to achieve interactive frame rates on mobile devices \cite{Reiser2024SIGGRAPH}.
While recent surface-based methods are also capable of representing thin structures like individual strands of grass \cite{Reiser2024SIGGRAPH, Yu2024TOG}, they still lag behind volume-based methods, especially when it comes to reconstructing highly fuzzy geometry like hair or plush.
Even if possible from a reconstruction perspective, a purely surface-based representation might be too memory-inefficient for representing extremely fuzzy objects \cite{Bhokare2024SIGGRAPH}.
Towards our goal of real-time view synthesis of fuzzy objects on mobile devices, we therefore focus on finding a more efficient volumetric formulation.
A key factor in the efficiency of volumetric representations is the number of samples required along a ray. State-of-the-art methods, such as SMERF and 3DGS, often require tens or even hundreds of samples per ray.
Additionally, the intrinsic characteristics of the rendering algorithm impact performance. In volume rendering, traversing an extra data structure to skip empty space increases memory bandwidth usage and results in suboptimal thread coherence.
For splatting, primitives must be sorted by their distance from the camera, a task that is challenging to implement efficiently on platforms with limited GPGPU capabilities.
For these reasons, such approaches are not suited for current mobile hardware, e.g. low-cost smartphones.
To address these limitations, we propose a differentiable representation that bounds the number of sampling points per ray to a small range (three to nine) and is sorting-free. This allows us, unlike 3DGS or SMERF, to achieve real-time rendering on low-cost smartphones.

Textured shells \cite{lengyel2000real, lengyel2001real, bears2014practical} have long been used in computer graphics to simulate fuzzy surfaces while minimizing geometric complexity. These are modeled as concentric, \emph{uniformly spaced}, semi-transparent layers, and are still widely used in modern game production.
Inspired by these techniques, our representation learns \emph{adaptively spaced} layers around a reconstructed object via gradient-based optimization. This allows for rasterization in a fixed order, from outermost to innermost, without the need for expensive empty space skipping or sorting during rendering.
However, it is non-trivial to learn the optimal spacing between individual layers; we tackle this by representing them as separate signed distance functions (SDFs) during training.
Before training all layers, we start by training a single opaque SDF that prevents degenerate solutions.
We then add the additional layers, constraining them from intersecting one another.
To further increase the expressivity of our representation, while keeping a low number of layers, we make each layer's transparency depend on the viewing direction.
All layers are optimized toward smooth solutions so that each can be baked into a lightweight mesh for efficient hardware-accelerated rasterization.
The simplicity of our meshes enables computing a high-quality UV parameterization, which is often problematic for highly complex, monolithic meshes \cite{Reiser2024SIGGRAPH}.
Finally, we optimize UV textures of spherical harmonics (SH) coefficients on the fixed geometry defined by our meshes.

We demonstrate how our approach renders significantly faster than volume-based and splatting-based methods, enabling high frame rates on a wide range of commodity devices, while simultaneously being more capable at representing fuzzy objects than single-surface approaches.

%

\section{Related Work}

\boldparagraph{Real-Time View Synthesis}
\label{sec:fast_rendering}
Neural radiance fields (NeRFs) achieved an impressive leap of quality by fitting a 3D scene representation via differentiable volume rendering to multi-view images \cite{Mildenhall2020ECCV}.
NeRFs represent the scene implicitly as a multi-layer perceptron (MLP) \cite{Mescheder2019CVPR, Park2019CVPR, Chen2019CVPR}.
Evaluating an MLP is arithmetically intensive, leading to slow inference.
To overcome this, a number of works explore faster representations based on 3D grids \cite{Liu2020NEURIPS, Reiser2021ICCV, Hedman2021ICCV, Yu2021ICCV, Garbin2021ICCV, esposito2022kiloneus}, triplanes \cite{Chen2022ECCV, Reiser2023SIGGRAPH, Duckworth2023TOG}, hash grids \cite{Mueller2022SIGGRAPH, takikawa2023compact}, or explicit primitives \cite{Aliev2020ECCV, Xu2022CVPRb, ruckert2022adop, chen2023neurbf, Kerbl2023SIGGRAPH}.
3D Gaussians have gained traction lately due to their fast training and rendering \cite{Kerbl2023SIGGRAPH}.
While these representations enhanced efficiency, rendering each pixel may still require a large number of samples per ray.

DONeRF reduces sample count by only sampling around depth values predicted by a separate MLP, which, however, requires access to ground-truth depth maps \cite{Neff2021CGF}.
Ada\-NeRF explicitly introduces sampling sparsity during the course of training \cite{kurz-adanerf2022}.
HybridNeRF introduces a hybrid surface–volume representation that encourages surfaces over volumetric rendering \cite{HybridNeRF}.
AdaptiveShells restrict sampling to a small shell around the surface \cite{Wang2020SIGGRAPHASIA}.
This shell is represented by a triangle mesh, rasterized to find the sampling range. Unlike AdaptiveShells, which uses volume rendering, our method finds all sampling locations via rasterization, enabling the use of 2D textures instead of 3D volume textures. Additionally, while AdaptiveShells learns a single SDF with a spatially-varying kernel, we learn multiple SDFs.
NDRF \cite{wan2023learning} extracts two mesh layers via marching cubes with varying density thresholds and rasterizes them to feature vectors that are decoded by a convolutional neural network (CNN). In turn, we store view-dependent opacities and colors as SH textures.
Similar to us, QuadFields also aggregate opacities and view-dependent colors from multiple ray-triangle intersections \cite{Sharma2024ECCV}.
However, these methods require significantly larger meshes, intersection calculations for all triangles along a ray, and a costly sorting process. In contrast, our formulation enables blending in a fixed order.

Another line of work aims for a single sample per ray.
MobileNeRF~\cite{chen2023mobilenerf} represents the scene with a coarse proxy mesh equipped with a binary opacity mask to increase expressivity.
BakedSDF~\cite{Yariv2023ARXIV}, NeRF2Mesh~\cite{Tang2023DelicateTM}, NeRFMeshing~\cite{rakotosaona2023nerfmeshing} and BOG~\cite{Reiser2024SIGGRAPH} first perform a 3D reconstruction of the scene, convert their respective training-time representation into a mesh, and then fit an appearance model to the mesh.
While these single-surface approaches are faster, they struggle to reconstruct fuzzy geometries. Even with accurate reconstruction from multi-view data, representing individual hairs explicitly would require substantial memory \cite{Bhokare2024SIGGRAPH}.

%
Gaussian frosting \cite{guedon2024frosting} anchors Gaussians on a tight surface shell to enhance editability. Like our approach, GaussianShellMaps \cite{abdal2024CVPR} utilize a multi-layer mesh representation. However, unlike our approach, GaussianShellMaps uses fixed shells rather than adaptively learning them, and predicts per-shell Gaussians. Further these methods still fall short of achieving real-time rendering on budget smartphones due to their reliance on splatting techniques.


\boldparagraph{3D Reconstruction}
Earlier methods for 3D reconstruction were often based on image matching \cite{Furukawa2010CVPR, Schoenberger2016CVPRa}.
More recent works directly fit level-set representations via differentiable rendering \cite{Niemeyer2020CVPR, Yariv2020NEURIPS, Oechsle2021ICCV, Vicini2022SIGGRAPH}.
To improve convergence, many approaches convert an SDF to a volumetric density on-the-fly, which is then used for standard volume rendering \cite{Yariv2021NEURIPS, Wang2021NIPS, neus2, Yu2022SDFStudio, Rosu2023CVPR, li2023neuralangelo}.
Recent methods demonstrate converting 2D or 3D Gaussians into high-quality surface representations, either through volumetric fusion \cite{Huang2DGS2024, Dai2024GaussianSurfels, Yu2024TOG, zhang2024ARXIV}, or using a Poisson reconstruction algorithm \cite{guedon2023sugar}.

\section{Preliminaries}
\label{sec:preliminary}

In this section, we provide a brief introduction to volume- and surface-based representations and related notations.

\boldparagraph{Rendering Equation}
A ray $\mathbf{r}$ in 3D space is parameterized by its origin \( \mathbf{o} \in \mathbb{R}^3 \) and unit direction \( \mathbf{v} \in \mathbb{S}^2 \).
A 3D point at distance $t$ along ray $\mathbf{r}$ is given by \( \mathbf{r}(t) = \mathbf{o} + \mathbf{v} t \).
Let $\sigma$ denote a density field, which assigns a nonnegative density value $\sigma(\mathbf x)\in\mathbb R_+$ to each 3D point $\mathbf x\in\mathbb R^3$, and let $\boldsymbol\xi$ be a vector field providing an RGB color $\xi(\mathbf x,\mathbf v)$ for each point $\mathbf x\in\mathbb R^3$ and direction $\mathbf v\in\mathbb S^2$.
We can render $\boldsymbol\xi$ along ray $\mathbf{r}$ for a given density field $\sigma$ using the following equation \cite{Mildenhall2020ECCV}:
\begin{equation}
   \mathcal{R}(\mathbf{r} \mid \sigma, \boldsymbol\xi) = \int_0^\infty \sigma(\mathbf r(t)) \, \boldsymbol\xi(\mathbf r(t), \mathbf v) \, w_\text{r}(t) \, dt \, \text{,}
   \label{eq:rendering}
\end{equation}
where $w_\text{r}(t)= \exp\!\left[-\int_0^t\sigma(\mathbf r(s)) \, ds\right]$.

\begin{figure}
   \centering
       %
       %
       %
       %
   \begin{subfigure}[b]{0.27\columnwidth}
       \centering \includegraphics[width=\columnwidth]{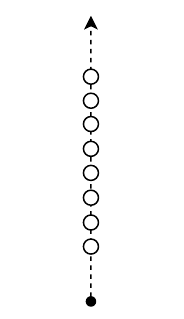}
       \caption{Volumetric}
       \label{fig:image_sampling_1}
   \end{subfigure}
   \hfill
   \begin{subfigure}[b]{0.27\columnwidth}
       \centering \includegraphics[width=\columnwidth]{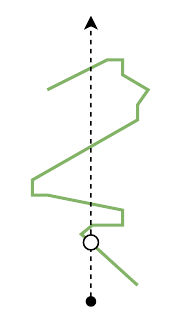}
       \caption{Surface}
       \label{fig:image_sampling_2}
   \end{subfigure}
   \hfill
   \begin{subfigure}[b]{0.27\columnwidth}
       \centering \includegraphics[width=\columnwidth]{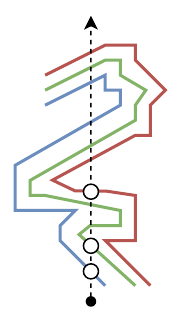}
       \caption{Ours}
       \label{fig:image_sampling_4}
   \end{subfigure}
   \caption{Sampling strategies: ($a$) volumetric rendering's dense sampling; ($b$) single sampling point, as in surface rendering; ($c$) our method, only sampling the first intersection with each surface.
   }
   \label{fig:sampling}
\end{figure}

\boldparagraph{Volume-Based Representations (NeRF)}
NeRF models a scene as a volume -- with absorption and emission but without scattering effects \cite{tagliasacchi2022volume} -- parametrized as $ (\sigma, \boldsymbol\xi) $.
NeRF numerically approximates \cref{eq:rendering} with quadrature \cite{max1995optical}, densely sampling rays to render each pixel (\cref{fig:image_sampling_1}).


\boldparagraph{Surface-Based Representations (NeuS)}
Many surface-based representations have been proposed \cite{Yariv2020NEURIPS, Mescheder2019CVPR, Yariv2021NEURIPS}; our work is built upon NeuS \cite{Wang2021NIPS}, as its densities weighting function -- for which we refer to the original paper -- is unbiased and occlusion-aware.
In short, NeuS represents a surface implicitly by modeling it through a neural SDF, trained with differentiable volumetric rendering.
A logistic distribution function $\phi_\beta$ maps distance values $d$ to densities as follows: 
\begin{equation}
   \phi_\beta(d) = \beta e^{-\beta d} / \left(1+e^{-\beta d}\right)^2\,.
   \label{eq:logistic}
\end{equation}
%
Here, the spread of densities around the surface (zero-level set) is controlled by the scalar $\beta$.
A small $\beta$ results in fuzzy densities, while in the limit $\beta \rightarrow \infty$ densities are sharp impulse functions for points on the implicit surface. 
While NeuS regards $\beta$ as a learnable parameter, \citet{Rosu2023CVPR} showed how controlling it explicitly leads to better reconstructions.
In our case, scheduling $\beta$ ensures that densities shift from being widely spread to being peaked by the end of the training.
When densities are peaked, the representation can be baked into a mesh suitable for efficient rendering.
However, in this case, all appearance information is condensed on a single point (\cref{fig:image_sampling_2}); therefore, surface-based methods are not able to model semi-transparent surfaces, strongly limiting their ability to handle mixed pixels often representing thin structures, which are notoriously missed \cite{Yariv2023ARXIV}.


\section{Method}
\label{sec:method}

In this section, we describe our proposed representation, Volumetric Surfaces, in detail. We begin with its implicit geometry definition, then cover rendering, training, and baking. We also discuss and justify our design choices.
Comprehensive architectural details can be found in the supplement.

\begin{figure*}[t]
    \centering
    \begin{subfigure}[b]{0.66\textwidth}
        \centering \includegraphics[width=\columnwidth, keepaspectratio, trim={0.4cm 0cm 0cm 0cm}]{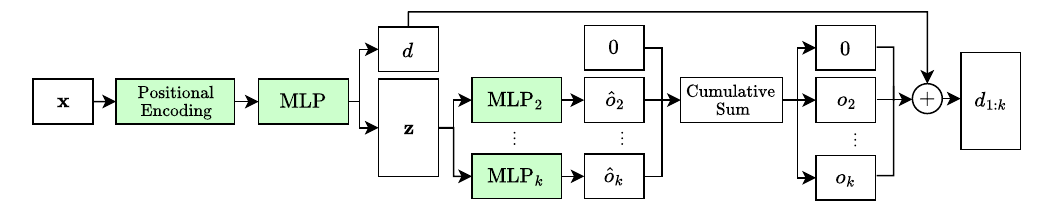}
        \caption{ }
        \label{fig:network}
    \end{subfigure}
    \hfill
    \begin{subfigure}[b]{0.33\textwidth}
        \centering \includegraphics[width=\columnwidth]{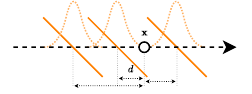}
        \caption{ }
        \label{fig:sdfs}
    \end{subfigure}
    \caption{
    ($a$) High-level architecture of our $k$-SDF network, predicting $k$ distance values as described in \cref{sec:k-sdf}. For simplicity of visualization, all offsets are positive.
    We highlight \colorbox{tabrealtime}{trainable} components.  
    For additional architectural details, refer to the supplementary material.
    ($b$) 1D example visualization of the output $d_{1:k}$ when evaluating the network at a sample point $\mathbf{x}$ along a ray. Signed distances are shown as solid lines, while $\beta$-controlled integration weights are represented as dotted lines.
    }
    \label{fig:geometry_net}
\end{figure*}

\subsection{\textit{k}-SDF}
\label{sec:k-sdf}
Our new representation, called $k$-SDF, models $k$ surfaces as distinct signed distance fields $\{d_1,\ldots,d_k\}$.
To composite the surfaces, $k$-SDF is endowed with an additional transparency field $\alpha$, which assigns a view-dependent transparency value $\alpha(\mathbf x,\mathbf v)\in[0,1]$ to each 3D point $\mathbf x$.
The transparency field enables modeling semi-transparent surfaces.
%
%
We generalize \cref{eq:rendering} to render a $k$-SDF $(d_{1:k}, \alpha, \boldsymbol\xi)$ as:
\begin{equation} 
    \mathcal R_\beta(\mathbf r \mid d_{1:k}, \alpha, \boldsymbol\xi) = \sum_{i=1}^{k}\mathcal R_\beta(\mathbf r \mid d_i, \boldsymbol\xi)\mathcal R_\beta(\mathbf r \mid d_i, \alpha) w_i \text{,}
    \label{eq:render_ksdf}
\end{equation}
where $w_i = \prod_{j=1}^{i}\mathcal (1- \mathcal R_\beta(\mathbf r \mid d_j, \alpha))$
and, for the sake of compactness, $\mathcal R_\beta(\mathbf r \mid d_i, \gamma)$ stands for $\mathcal R(\mathbf r \mid \phi_\beta\circ d_i, \gamma)$.
Our per-surface density field is derived from the signed distance field $d_i$ as in \cref{eq:logistic}.

\begin{figure}
    \centering
    \renewcommand{\arraystretch}{0.5} 
    \setlength{\tabcolsep}{1pt}     
    \begin{tabular}{%
        >{\centering\arraybackslash}m{0.05\columnwidth}
        >{\centering\arraybackslash}m{0.95\columnwidth}
    }
        $a$) &
        \includegraphics[width=0.65\columnwidth]{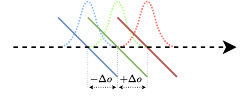} \\
        $b$) &
        \includegraphics[width=0.65\columnwidth]{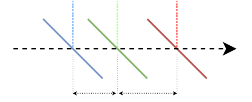} \\
    \end{tabular}
    \caption{1D example visualization of 3-SDF along a ray. Signed distances are shown as solid lines, while $\beta$-controlled integration weights are represented as dotted lines.
    ($a$) Initialization of the support surfaces as positive and negative constant displacements $\Delta o$ from the main SDF.
    ($b$) Densities peaked at the end of training.
    The two support surfaces are displace by trained offsets $(o_1, o_2)$, $d$: $ \mathcolor{surfblue}{d_1} = \mathcolor{surfgreen}{d} - o_1$, $\mathcolor{surfred}{d_2} = \mathcolor{surfgreen}{d} + o_2$.}
    \label{fig:ray_weights}
\end{figure}

\boldparagraph{Support Surfaces as Shells} 
Blending weights in \cref{eq:render_ksdf} are order-dependent, but lower-ranked densities are not necessarily positioned closer to the camera.
To avoid sorting, we model the set of surfaces as shells, ensuring that layers are traversed in ray-intersection order.
We model our $k$-SDF with a main surface represented as an SDF $d$, and a set of $k\!-\!1$ support surfaces represented as offset fields $\{o_2,\ldots,o_k\}$ (\cref{fig:ray_weights}) from $d$'s zero-level set.
The signed distance functions for each surface are thus given by: $d_{1:k} = (d, d+o_2,\ldots, d + o_k)$.
A surface defined by a positive offset is \emph{contained} inside the main one, while a negative offset yields a surface \emph{containing} the main one.
To model multiple support surfaces while enforcing order, we perform a cumulative sum over predicted relative offsets $\{\hat{o}_2,\ldots,\hat{o}_k\}$ (separately for negative and positive offsets) and use the resulting absolute displacements $\{o_2,\ldots,o_k\}$ to parameterize the surfaces.
\cref{fig:geometry_net} illustrates our model.

\boldparagraph{Rendering Individual Surfaces}
\cref{eq:rendering} is approximated with quadrature~\cite{Mildenhall2020ECCV} and evaluated at $n$ discrete points $\{\mathbf x_1, \ldots, \mathbf x_n\}$ sampled along a ray $\mathbf{r}$.
For surface $j$, we render its color $\mathcal C_j(\mathbf r)$ and transparency $\mathcal A_j(\mathbf r)$ as:
\begin{align}
    \mathcal{C}_j(\mathbf{r}) &= \sum_{i=1}^n w_{i, j} \boldsymbol\xi(\mathbf{x}_i, \mathbf{v}, \mathbf{n}_{i, j}, \mathbf{z}_{i}) \text{,} \\
    \mathcal{A}_j(\mathbf{r}) &= \sum_{i=1}^n w_{i, j} \alpha(\mathbf{x}_i, \mathbf{v}, \mathbf{n}_{i, j}, \mathbf{z}_{i}) \text{,}
\end{align}
where RGB and transparency fields ($\boldsymbol\xi$, $\alpha$) are conditioned on sample position $\mathbf{x}_i$, view direction $\mathbf{v}$, SDF normal $\mathbf{n}_{i,j}$, and feature vector $\mathbf{z}_i$, an additional output of the $k$-SDF model. Sample weights $w_{i, j}$ are computed as in \citet{Wang2021NIPS}.



\subsection{Surfaces Blending}
\label{sec:surfaces_blending}

We now rewrite \cref{eq:render_ksdf} as a fixed-order alpha blending of the rendered surface color and transparency:
\begin{equation}
    \mathcal{R}(\mathbf{r}) = \sum_{i=1}^{k} \mathcal{C}_i(\mathbf{r}) \mathcal{A}_i(\mathbf{r}) w_i \text{,} 
\end{equation}
where $w_i = \prod_{j=1}^{i} \left( 1 - \mathcal{A}_{j}(\mathbf{r}) \right)$.

\boldparagraph{Transparency Attenuation}
Blending different hard surfaces might result in clear cut-offs at their borders, especially noticeable from test views.
We prefer smoother transitions towards full transparency.
We achieve this by multiplying predicted transparencies with a weight $\alpha_{\text{w}}$ that depends on the angle between the view direction and the surface normal, downweighting the contribution from grazing angles. 
Specifically, we use: $\alpha_\text{w} = 2 \cdot \text{Sigmoid} (10 \cdot |\mathbf{v} \cdot \mathbf{n}|) - 1$.


\subsection{Training and Baking}

Our method is composed of two main stages.
First, we optimize an implicit representation of $k$ surfaces and their appearances.
Then, we fine-tune resolution-bounded neural textures over their explicit (meshes) representation.
Our hyperparameters are robust, ensuring consistent performance across all scenes and datasets.
In the following, we describe both phases in detail.

\boldparagraph{Implicit Surfaces}
We begin by training a standard NeuS-like~\cite{Wang2021NIPS} model for 100k iterations, during which $\beta$ is exponentially scheduled from large densities $\phi_{\beta_1}$ to thinner densities $\phi_{\beta_2}$.
%
In practice, training the $k$-SDF model from scratch with predicted transparencies often introduces fully transparent additional geometry in the reconstruction. This opaque pre-training step helps prevent such artifacts.
The reconstructed surface serves as an anchor for initializing the remaining $k \!-\! 1$ support surfaces, which are represented as shells uniformly spaced from the main surface by $\Delta o$.
The mathematical formulation of $k$-SDF (\cref{sec:k-sdf}) allows for initializing support surfaces both inside and outside the main SDF.
However, we find that initializing all support surfaces on the inside increases model capacity, leading to better results (\cref{sec:ablations}).
%

While we have observed the optimization to be robust under different values of $\Delta o$, we found a good balance by setting it by the logistic distribution function $\phi_{\beta_{2}}$ standard deviation:  $\Delta o = (1/\beta_{2})\pi / \sqrt{3}$. This ensures that surface densities only partially overlap (e.g., \cref{fig:ray_weights}).
The $k$-SDF model is trained for 50k iterations starting from $\phi_{\beta_2}$, until a $\phi_{\beta_3}$ value for which all surfaces are modeled as peaked densities.
When this happens, our rendering process collapses to a simple blending of hard surfaces (\cref{fig:image_sampling_4}), making the reconstructed set of implicit surfaces optimal for the later steps.
During both training phases, we apply two additional losses to all surfaces.
The Eikonal loss $\mathcal{L}_\text{e}$~\cite{Gropp2020ICML}, calculated on points in the vicinity of the zero-level sets and on randomly sampled points, and a curvature loss $\mathcal{L}_\text{s}$~\cite{Rosu2023CVPR} on near-surfaces points, to push the optimization toward smooth solutions.
We minimize $\mathcal{L} = \mathcal{L}_\text{c} +\lambda_\text{e} \mathcal{L}_\text{e} + \lambda_\text{s} \mathcal{L}_\text{s}$, where $\mathcal{L}_\text{c}$ is the standard $L_1$ pixel-wise color loss, $\lambda_\text{e} = 0.04$ and $\lambda_\text{s} = 0.65$.

\boldparagraph{Occupancy Grid}
As implicit surfaces training progresses and densities peak, our volumetric rendering samples are gradually positioned closer to the zero-level sets of the signed distance functions, since points farther away would be in empty space.
To do so, we compute per-surface binary occupancy values by describing as occupied voxels whose center point's $\left| d \right|$ value, together with the current $\beta$ and the voxel's space diagonal, would allow any point in the voxel volume to have density above a $10^{-4}$ threshold.
Per-surface occupancy values are then aggregated with an \texttt{or} operation, resulting in a single binary occupancy grid.
Rays traverse the grid to sample $n$ points uniformly in occupied space.
In our experiments, we used a grid of resolution $256^3$.

\boldparagraph{Importance Sampling}
During implicit surfaces training, we adopt hierarchical sampling from NeuS \cite{Wang2021NIPS}, extending it to the $k$-surfaces case.
Starting from the $n$ uniform samples, each SDF is evaluated to compute $k$ CDFs; these are summed together and normalized.
The resulting probability distribution is used to sample $m/2$ additional points that are added to the previous $n$.
The whole operation is then repeated on the expanded set of points to add $m/2$ more samples.
The resulting number of samples per ray is then $n + m$.
In the first iteration, the kernel size $\beta$ is half of that used during rendering, while in the second, it matches it.

\boldparagraph{Baking Meshes}
After the implicit surfaces optimization, we extract $k$-SDF zero-level sets as high-resolution meshes using marching cubes \cite{Lorensen1987SIGGRAPH} and simplify them significantly \cite{Garland1997SIGGRAPH, maggioli2023rematching} to 0.02\% of the original triangle count (approximately 2\,MB per mesh for synthetic scenes), meeting the strict computing and memory constraints of mobile hardware.
%
%
Finally, we generate UV atlases for these lightweight meshes using \verb|xatlas|~\cite{young2020xatlas}.
%
This step is essential for training per-mesh neural SH textures, which are later baked into explicit textures.
%



\begin{figure}[b]
    \centering
    \includegraphics[width=\columnwidth]{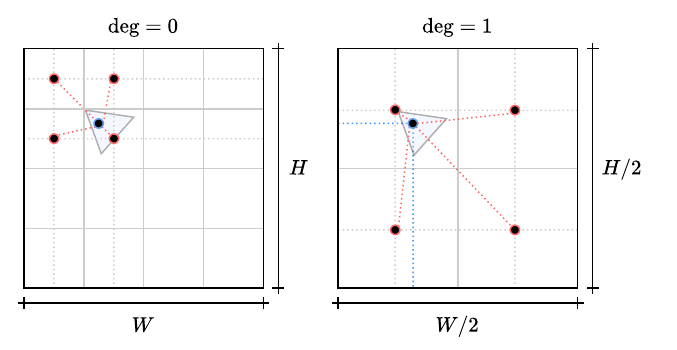}
    \caption{Bilinear interpolation in our mixed-resolution textures.
    Instead of querying the blue 2D point directly, we predict the values at its surrounding texel centers (red points) and bilinearly interpolate them. This anchors the neural texture to a predefined target resolution ($W$, $H$).}
    \label{fig:neural_textures}
\end{figure}

\boldparagraph{Texturing Meshes}
\label{sec:meshes_texturing}
Finally, we train a per-surface view-dependent appearance model for color and transparency using neural textures.
A neural texture is implemented as a hash-grid with input dimension 2 and a small decoder MLP with output dimension dependent on the SH degree it models.
The $i$-th neural texture predicts $\overline{\mathbf{sh}_i} \in \mathbb{R}^{4 \times c_{\text{deg}}}$, with $c = \{1, 3, 5, 7\}$.
The outputs of all neural textures are concatenated over the coefficient dimension ($\mathbf{sh} \!\in\! \mathbb{R}^{4 \times (\text{deg} + 1)^2} $) and decoded with view direction $\mathbf{v}$ to predict RGBA.
Specifically, we discretize our neural textures by grounding them to a fixed target baking resolution.
In this setting, each point $\mathbf{x} \!\in\! \mathbb{R}^3$ on the surface of a mesh is mapped to a point $\overline{\mathbf{x}} \!\in\! [0, 1]^2$ in UV space.
SH coefficients at $\overline{\mathbf{x}}$ are predicted as the result of a bilinear interpolation of predictions at neighboring texel centers $\{ \overline{\mathbf{x}}_{i} \}_{1, \ldots, 4}$, as visualized in \cref{fig:neural_textures}.
This effectively mimics how OpenGL fragment shaders access and interpolate texture values; by doing so, we optimize our texture values such that when displayed in our real-time renderer, images match -- up to numerical precision -- those synthesized by our differentiable renderer.
This final phase is trained for 15k iterations using the $L_1$ pixel-wise color loss.

\begin{table}[t]
    \caption{\label{tab:textures_res}%
        Sweep over the number of meshes and target neural textures resolution.
        Mixed resolutions are described in \cref{sec:meshes_texturing}.
        Increasing neural texture resolution improves reconstruction quality up to $1024^2$, but performance declines beyond that.
        Modeling all textures at the maximum resolution ($2048^2$) is counterproductive, as results indicate that a mixed-resolution approach yields better image quality while reducing memory usage.
        Results on the \textit{khady} scene from Shelly \cite{Wang2020SIGGRAPHASIA}; we highlight the \colorbox{tabfirst}{best}, \colorbox{tabsecond}{second best} and \colorbox{tabthird}{third best}.
        }
    \centering
    \resizebox{0.85\columnwidth}{!}{
    \begin{tabular}{l|ccccc}
    \toprule
    \multicolumn{1}{c}{} & \multicolumn{5}{c}{PSNR $\uparrow$} \\ \cmidrule(rl){2-6}
    \textit{Method} & mixed & 2048\textsuperscript{2} & 1024\textsuperscript{2} & 512\textsuperscript{2} & 256\textsuperscript{2} \\
    \midrule
    5-Mesh & 
    \colorbox{tabsecond}{29.88} & \colorbox{tabsecond}{29.88} & \colorbox{tabfirst}{29.91} & \colorbox{tabthird}{29.82} & 29.54 \\ 
    7-Mesh & 
    \colorbox{tabsecond}{29.97} & \colorbox{tabsecond}{29.97} & \colorbox{tabfirst}{30.02} & \colorbox{tabthird}{29.94} & 29.65 \\ 
    9-Mesh & 
    \colorbox{tabsecond}{29.96} & \colorbox{tabthird}{29.93} & \colorbox{tabfirst}{29.97} & 28.91 & 29.62 \\ 
    \bottomrule
    \end{tabular}
    }
\end{table}

\boldparagraph{Mixed Resolutions}
Storing all textures at full resolution ($2048^2$) is impractical, as it would require approximately 0.5\,GB per mesh.
We propose scaling texture resolution proportionally to the spherical harmonics (SH) coefficient degree: the base color is modeled at the highest resolution ($2048^2$), while the highest SH degree coefficients are modeled at the lowest resolution ($256^2$).
This approach significantly reduces the memory footprint to approximately 14\,MB per mesh without compromising image quality (\cref{tab:textures_res}).

\boldparagraph{Squeezing and Quantization}
Neural textures predicted values are continuous and unbounded.
Before storing them, they must be squeezed to the unit range via $\hat{\mathbf{sh}} = \text{Sigmoid}(\mathbf{sh})$.
Training must account for quantization to the discrete $[0, 255]$ range when storing textures as images.
Following MERF \cite{Reiser2023SIGGRAPH}, we apply the function $q(x) = \text{round}(255x) / 255$ to the predicted and squeezed coefficients.
Before rendering, we re-scale value to a hyper-parameters controlled range $[v_\text{min}, v_\text{max}]$: $\mathbf{sh} = v_\text{min} + (v_\text{max} - v_\text{min}) q(\hat{\mathbf{sh}})$.
We observed a range of $[-15, 15]$ to work well on all scenes.

\boldparagraph{Baking Textures}
Lastly, we bake our resolution-grounded neural textures, which is straightforward as it only requires predicting values at texel centers and storing them.
Baking results in $(\text{deg}+1)^2$ PNG images, where the $i$-th image stores the $i$-th coefficient of RGBA channels.
The fully baked representation can finally be visualized in our WebGL renderer, which rasterizes all meshes in a fixed order and blends them in the final frame buffer before displaying it on screen.


\section{Evaluation}
\label{sec:evaluation}

\begin{table}
    \caption{
    Framerate is measured on \emph{close-up} views at HD (720p) resolution on low-power smartphone (Samsung A52s) (marked with $\diamond$) and laptop (Dell XPS 13 i5) (marked with $\star$), on respective WebGL renderers; memory footprint as stored on disk.
    3DGS~\cite{Kerbl2023SIGGRAPH} with spherical harmonics of degree 2, ours of degree 3. 
    Metrics averaged over scenes of the Shelly~\cite{Wang2020SIGGRAPHASIA} dataset.
    We present further qualitative comparisons in the supplementary material.
    }
    \centering
    \resizebox{0.9\columnwidth}{!}{
    \begin{tabular}{l|rrrr}
    \toprule
    \textit{Method} & \textit{FPS $^\diamond \uparrow$} & \textit{FPS $^\star \uparrow$} & \textit{PSNR $\uparrow$} & \textit{MB $\downarrow$} \\
    \midrule
    MobileNeRF~\cite{chen2023mobilenerf} & 24 & 35 & 29.30 & 194 \\
    3DGS-50K & 20 & 160 & 32.73 & 12 \\
    3DGS-75K & 13 & 115 & 33.05 & 18 \\
    3DGS~\cite{Kerbl2023SIGGRAPH} & 8 & 18 & 35.44 & 57 \\
    \midrule
    3-Mesh & 65  & 145 & 33.39 & 46 \\
    5-Mesh & 55  & 90 & 34.25 & 77 \\
    7-Mesh & 42  & 70 & 34.50 & 110 \\
    9-Mesh & 35  & 55 & 34.38 & 140 \\
    \bottomrule
    \end{tabular}
    }
    \label{tab:performance}
\end{table}

\begin{table*}[!ht]
    \caption{\label{tab:main_results}%
    Results are averaged across all testing scenes; we highlight the \colorbox{tabfirst}{best}, \colorbox{tabsecond}{second best} and \colorbox{tabthird}{third best} results.
    Highlighted \colorbox{tabrealtime}{baselines} meet the compute and/or memory requirements to run on general-purpose hardware (via WebGL) without modifications, as discussed in \cref{sec:fast_rendering}.
    QuadFields~\cite{Sharma2024ECCV} fails to meet real-time requirements due to its reliance on specialized ray-tracing acceleration hardware and its high memory consumption (e.g., Shelly~\cite{Wang2020SIGGRAPHASIA} scenes require an average of 1213\,MB).
    \textbf{Our results are directly computed from our WebGL real-time render on our final baked representation.}
    Methods marked with a ``$\star$'' show results from original papers, as code is unavailable at the time of writing. 
    Instant-NGP~\cite{Mueller2022SIGGRAPH} results on Shelly from \citet{Wang2020SIGGRAPHASIA}.
    DTU~\cite{jensen2014large} is evaluated on masked foreground.
    %
    %
    PermutoSDF~\cite{Rosu2023CVPR} trained until densities are fully peaked ($\phi_{\beta_3})$.
    Metrics not provided by a baseline are denoted with “---”.
    Please note that the Shelly dataset, as released by the authors, has a large overlap of views between test and training sets. All our experiments were conducted on a cleaned version of the dataset, free of this problem. However, a fair comparison with baselines whose code is not public remains difficult.
    Our per-scene metrics are reported in the supplementary material.}
    \centering
    \resizebox{0.9\linewidth}{!}{
    \begin{tabular}{l|ccc|ccc|ccc}
        \toprule
        & \multicolumn{3}{c|}{Shelly~\cite{Wang2020SIGGRAPHASIA}} & \multicolumn{3}{c|}{Custom (\textit{plushy} + \cite{Sharma2024ECCV})} & \multicolumn{3}{c}{DTU~\cite{jensen2014large}  (\textit{scans 83}, \textit{105})} \\
        \midrule
        \textit{Method} & PSNR $\uparrow$ & SSIM $\uparrow$ & LPIPS $\downarrow$ & PSNR $\uparrow$ & SSIM $\uparrow$ & LPIPS $\downarrow$ & PSNR $\uparrow$ & SSIM $\uparrow$ & LPIPS $\downarrow$ \\ 
        \midrule
        \colorbox{tabrealtime}{3DGS~\cite{Kerbl2023SIGGRAPH}} & 
        \colorbox{tabsecond}{35.44} & 0.975 & \colorbox{tabthird}{0.089} & 
        \colorbox{tabfirst}{37.34}  & \colorbox{tabfirst}{0.982} & \colorbox{tabsecond}{0.147} &
        \colorbox{tabsecond}{38.06} & \colorbox{tabfirst}{0.989} & 0.086 \\
        Instant-NGP~\cite{Mueller2022SIGGRAPH} & 33.22 &  0.922 & 0.125 &
        31.13 &  0.935 & \colorbox{tabfirst}{0.132} & 
        \colorbox{tabfirst}{38.24} & --- & --- \\
        PermutoSDF~\cite{Rosu2023CVPR} & 
        29.85 & 0.950 &  0.129 &  
        33.31 &  0.961 & 0.193 &
        36.31 & \colorbox{tabsecond}{0.988} & 0.098 \\
        AdaptiveShells$^\star$~\cite{Wang2020SIGGRAPHASIA}  & 
        \colorbox{tabfirst}{36.02} & 0.954 & \colorbox{tabsecond}{0.079} & 
        --- & --- & --- &
        --- & --- & --- \\
        QuadFields$^\star$~\cite{Sharma2024ECCV}  &  
        \colorbox{tabthird}{35.13} & 0.954 & \colorbox{tabfirst}{0.073} & 
        --- & --- & --- &
        --- & --- & --- \\
        \colorbox{tabrealtime}{MobileNeRF~\cite{chen2023mobilenerf}}  &  
        29.30 & 0.939 &  0.150 &    
        30.89 & 0.942 & 0.195 &
        --- & --- & --- \\
        \midrule
        3-Mesh &  
        33.39 & 0.978 & 0.115 &  
        35.00 & 0.970 & 0.171 &
        36.41 & 0.985 & 0.092 \\
        5-Mesh &  
        34.25 & \colorbox{tabthird}{0.980} & 0.110 &
        35.45 & 0.975 & 0.172 &
        36.87 & 0.986 & \colorbox{tabthird}{0.085} \\
        7-Mesh &  
        34.50 & \colorbox{tabfirst}{0.981} & 0.109 &  
        \colorbox{tabthird}{35.63} & \colorbox{tabthird} {0.977} & 0.169 &  
        36.77 & \colorbox{tabthird}{0.987} & \colorbox{tabsecond}{0.084} \\
        9-Mesh &  
        34.38 & \colorbox{tabsecond}{0.981} & 0.110 & 
        \colorbox{tabsecond}{35.74} & \colorbox{tabsecond}{0.978} & \colorbox{tabthird}{0.167} &
        \colorbox{tabthird}{37.17} & 0.987 & \colorbox{tabfirst}{0.083} \\
        \bottomrule
        \end{tabular}
    }
\end{table*}

\begin{figure}
    \centering
    \resizebox{1.0\columnwidth}{!}{%
        \input{figures/pgf/tradeoffs.pgf}
        \hspace{1.5em}
    }%
    \caption{
      Frame rate vs. image quality comparison (smartphone results $\diamond$ from \cref{tab:performance}).
      The radius of each circle represents the memory footprint as stored on disk. The vertical dashed line marks the required frame rate for real-time rendering (30 FPS).}
    \label{fig:results_tradeoffs}
\end{figure}

\begin{figure*}[t]
    \centering
    \begin{tabular}{cccc} 
        \includegraphics[width=0.18\textwidth, keepaspectratio=false, trim=423 0 423 0, clip]{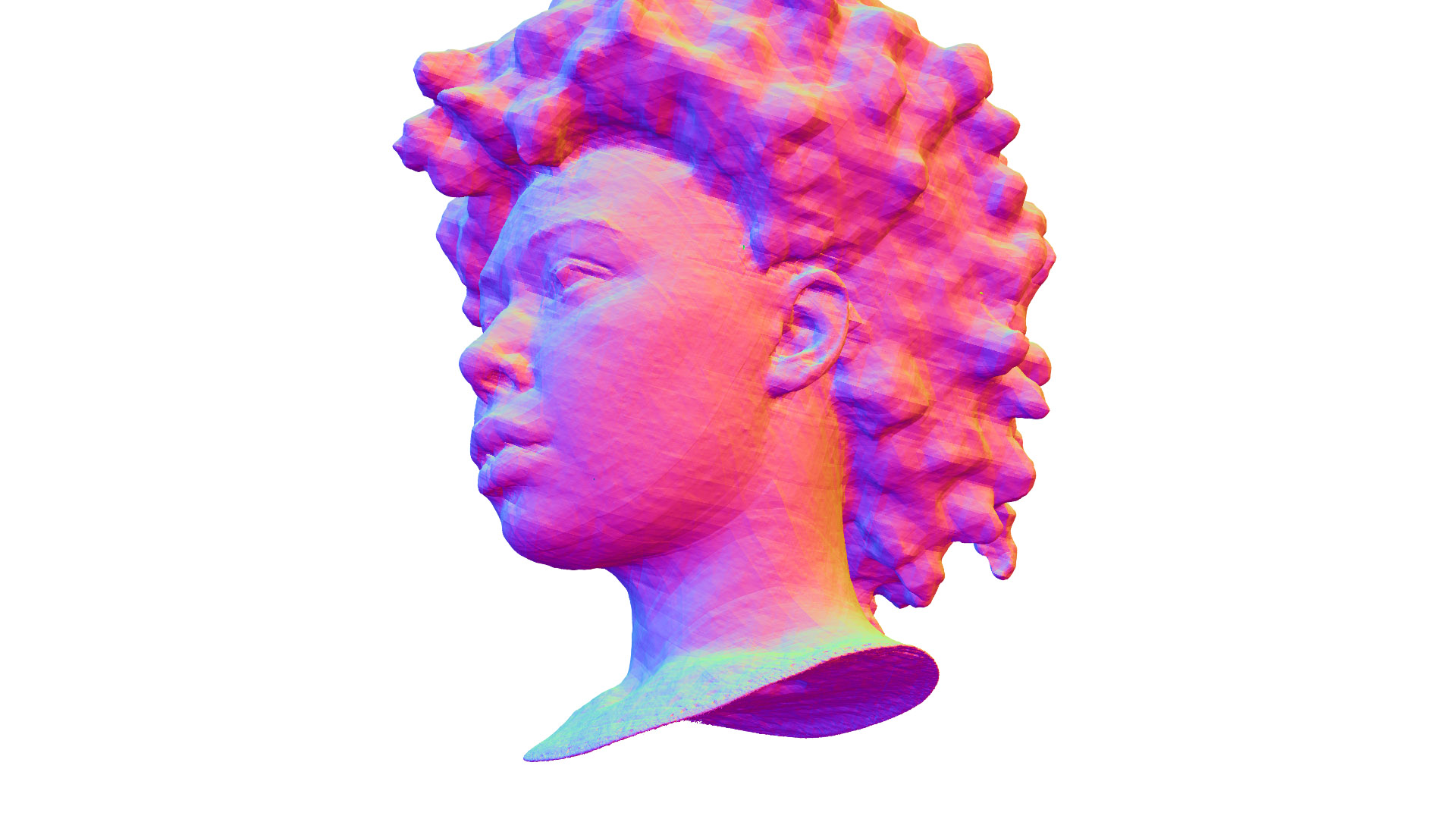} & 
        \includegraphics[width=0.18\textwidth]{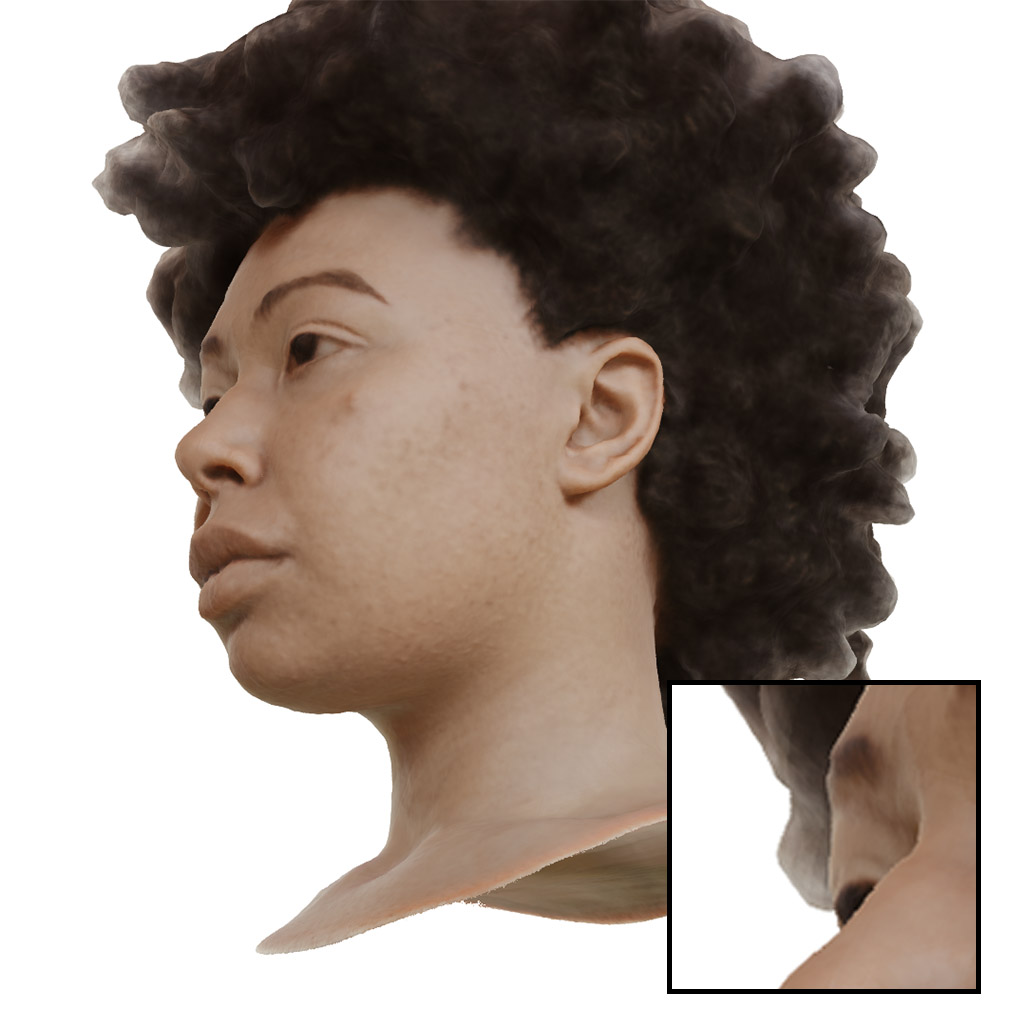} & 
        \includegraphics[width=0.18\textwidth ]{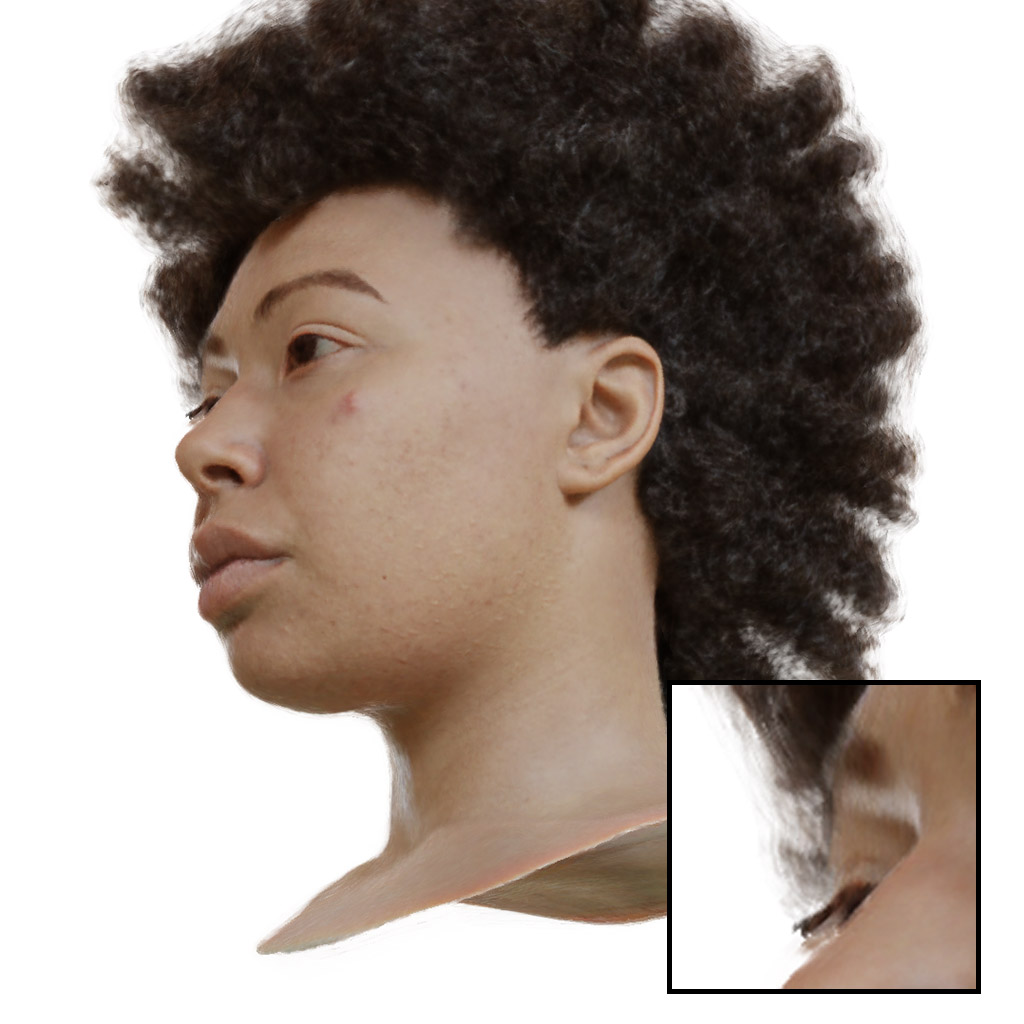} & 
        \includegraphics[width=0.18\textwidth]{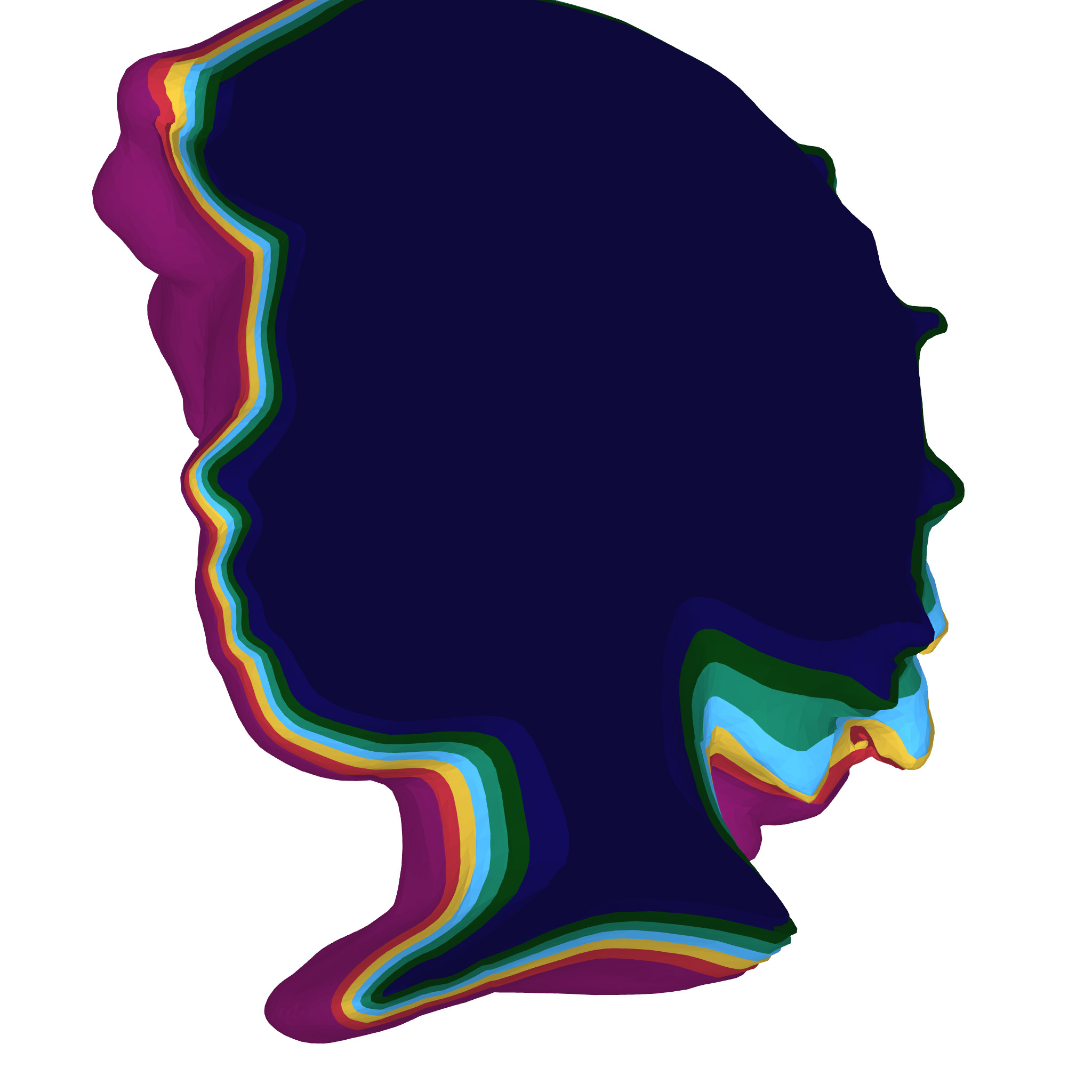} \\
        \includegraphics[width=0.18\textwidth, keepaspectratio=false, trim=423 0 423 0, clip]{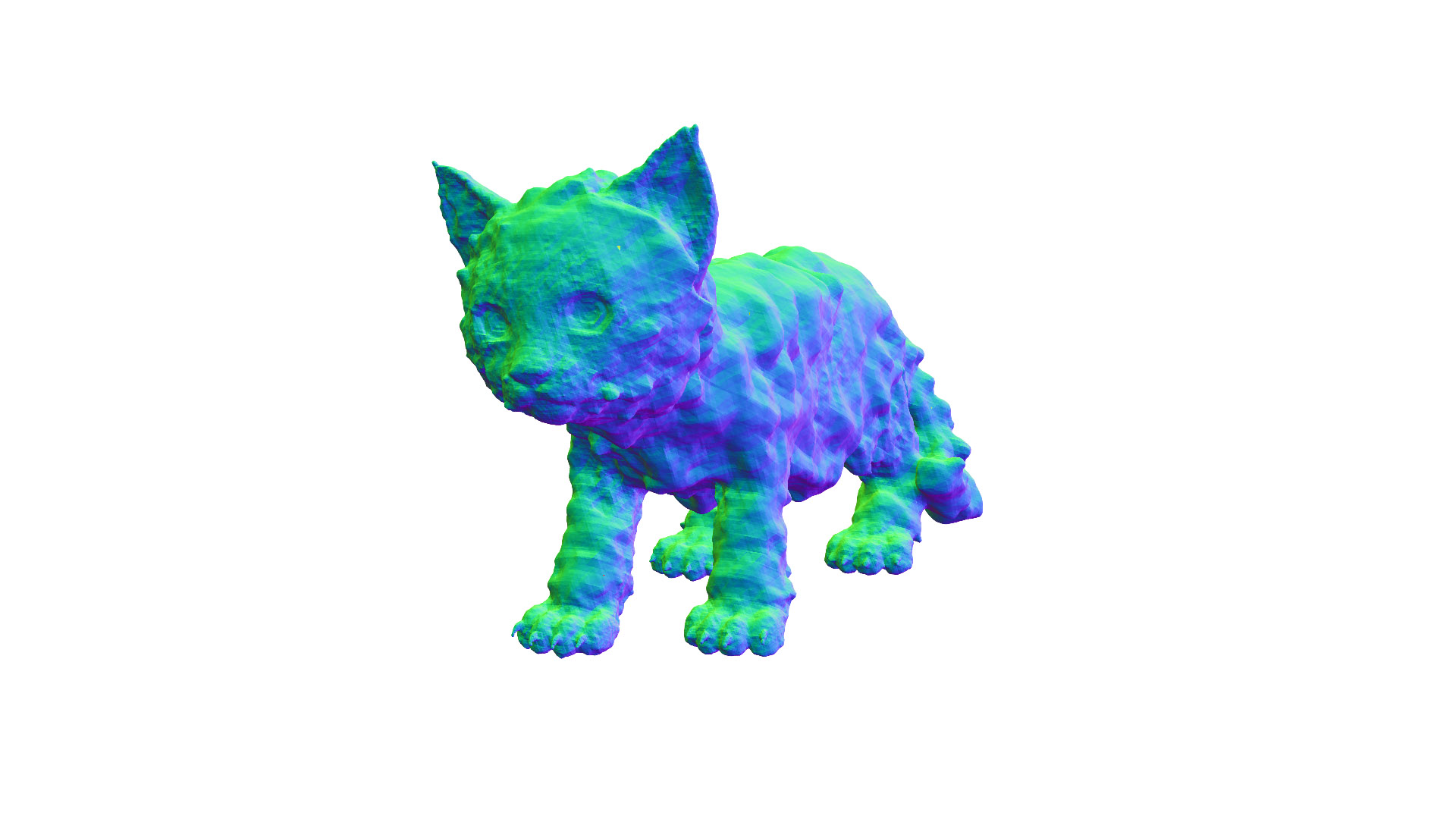} & 
        \includegraphics[width=0.18\textwidth]{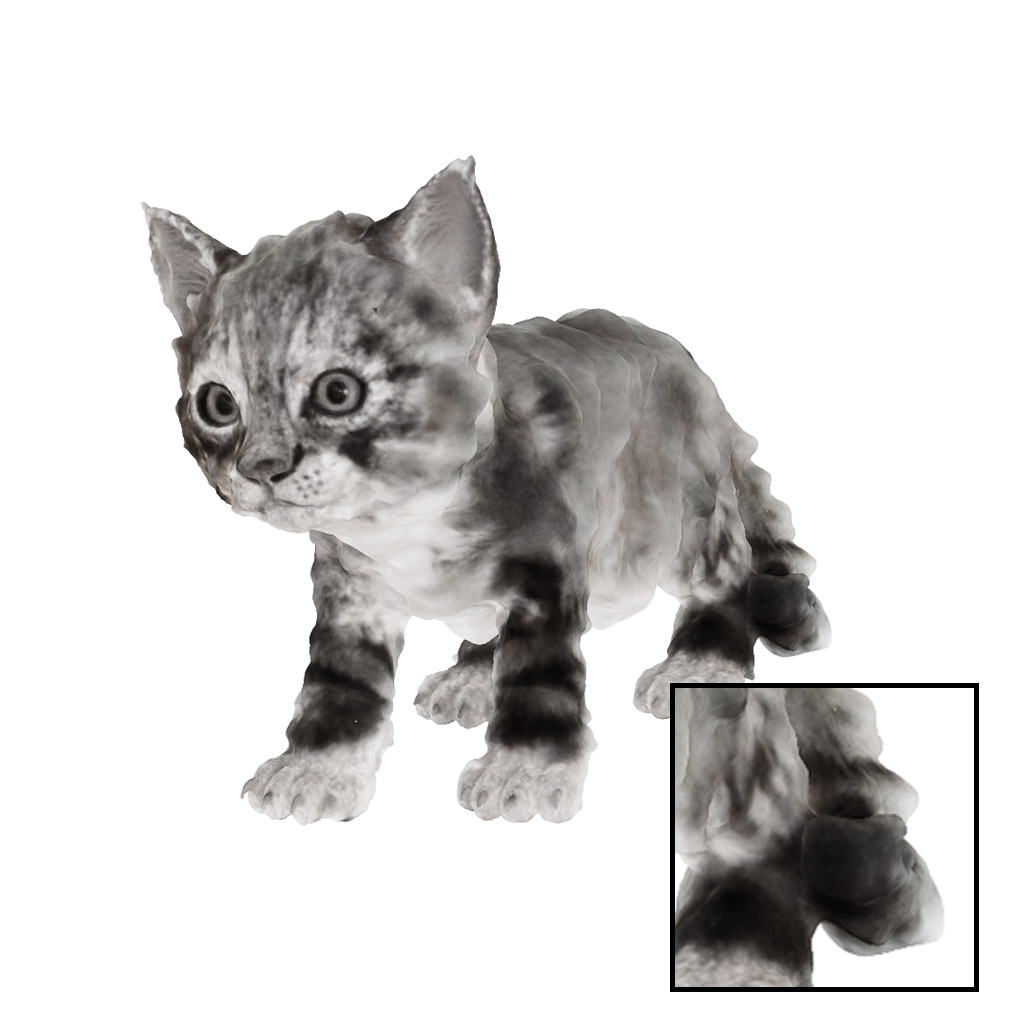} & 
        \includegraphics[width=0.18\textwidth]{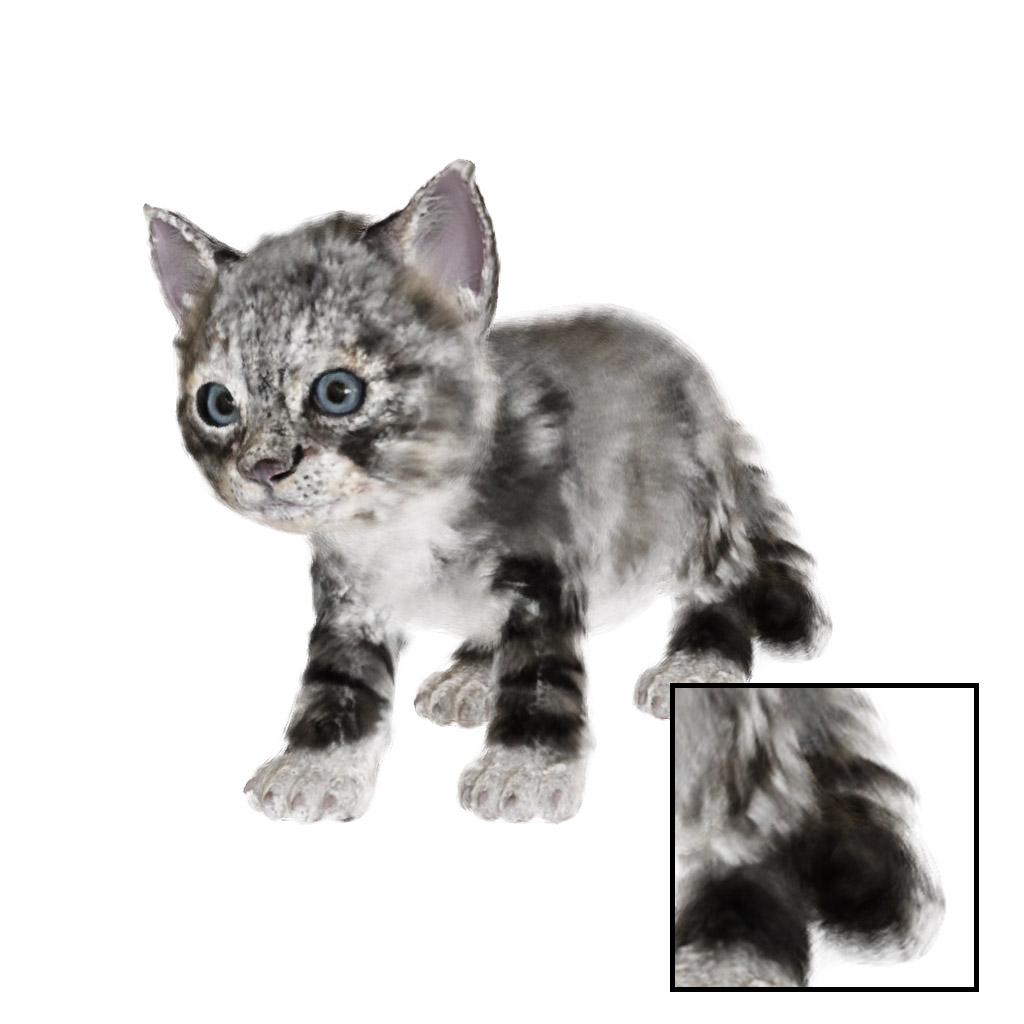} & 
        \includegraphics[width=0.18\textwidth]{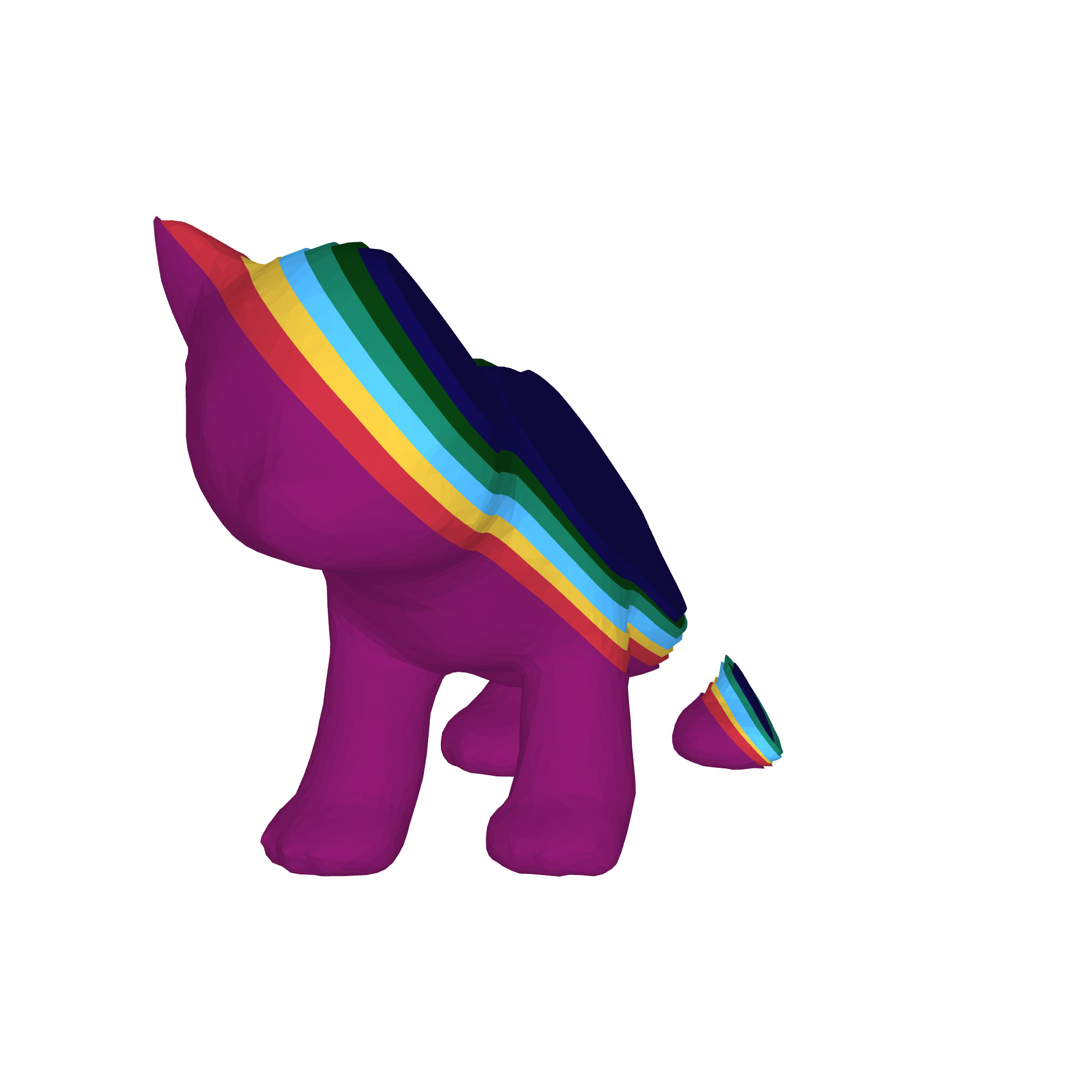} \\
        \includegraphics[width=0.18\textwidth]{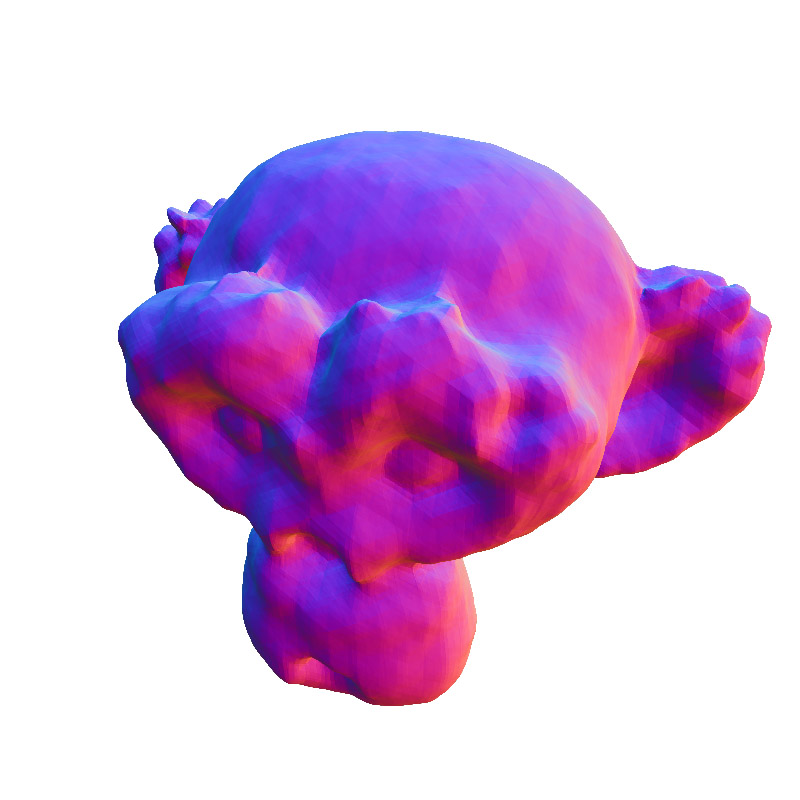} & 
        \includegraphics[width=0.18\textwidth]{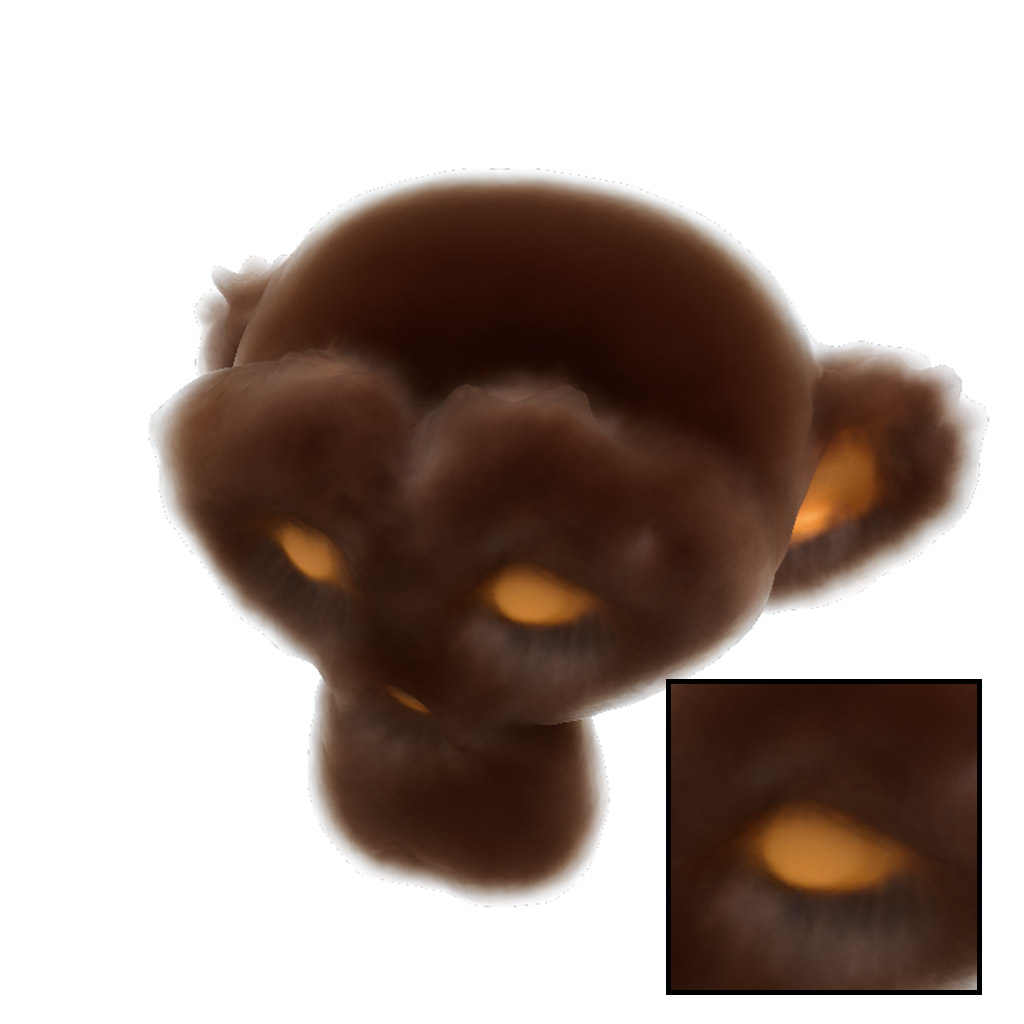} & 
        \includegraphics[width=0.18\textwidth]{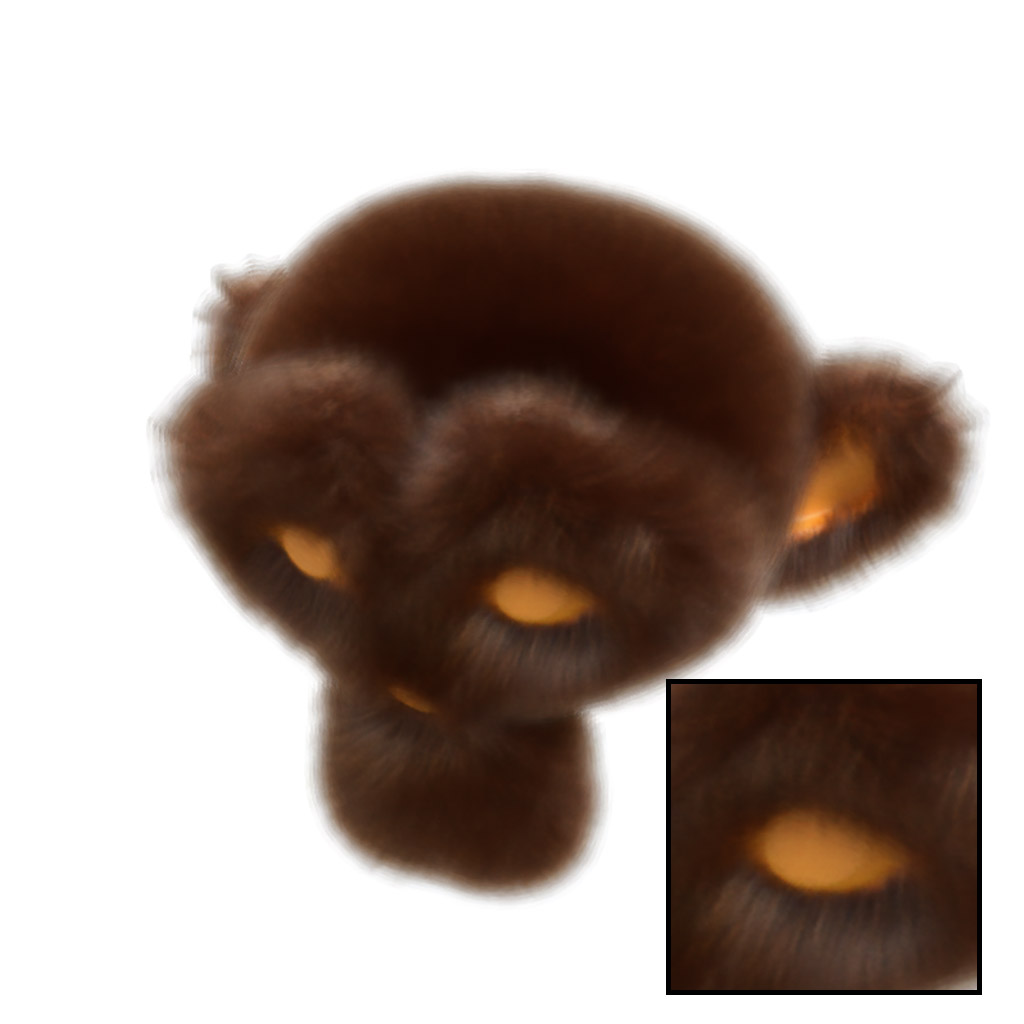} & 
        \includegraphics[width=0.18\textwidth]{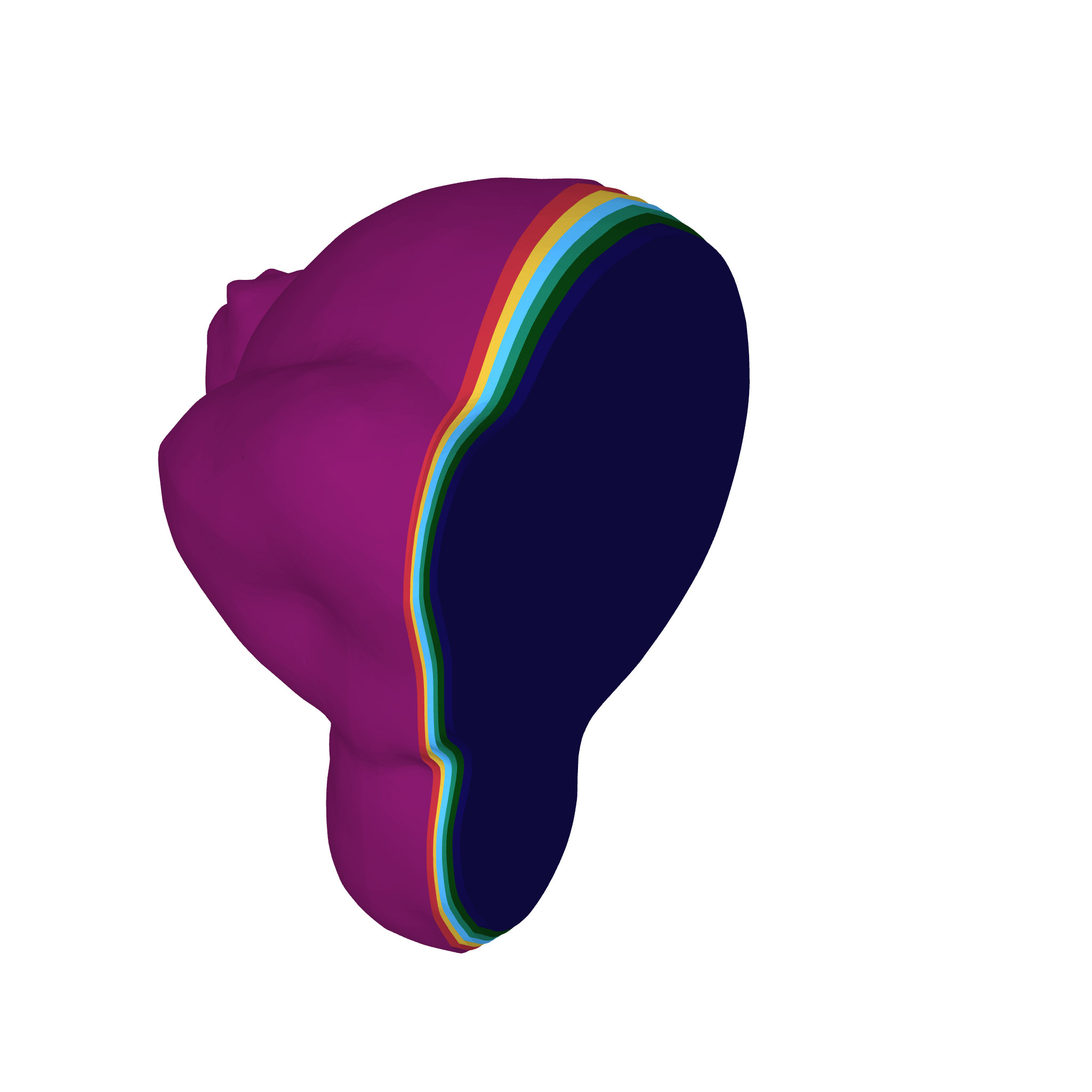} \\
        $a$) & $b$) & $c$) & $d$) 
    \end{tabular}
    \caption{
    Qualitative comparisons between ($b$) PermutoSDF~\cite{Rosu2023CVPR}, a state-of-the-art implicit surface-based method, and ($c$) our Volumetric Surfaces demonstrate that our approach convincingly represents fuzzy objects. This is achieved by trading ($a$) high-frequency geometry for the number of integration points, which are found by rasterizing smooth, lightweight meshes defined as ($d$) shells around the object and traversed in a fixed order. Volumetric Surfaces enable fast rendering of fuzzy geometries on general-purpose hardware with image quality approaching the latest volumetric representations (\cref{tab:main_results}).
    Scenes from Shelly~\cite{Wang2020SIGGRAPHASIA} and QuadFields~\cite{Sharma2024ECCV}.
    }
    \label{fig:qualitative}
\end{figure*}

Our key baselines focus on real-time rendering, with 3DGS and MobileNeRF as primary competitors. 3DGS represents the fastest volumetric approach, while MobileNeRF pioneers surface-based neural graphics for mobile devices. 
We compare to these methods in terms of quality, speed, and memory footprint on a low-power laptop (Dell XPS 13 i5) and a smartphone (Samsung A52s) (\cref{tab:performance}).
Additionally, we compare with other baselines not designed for general-purpose hardware rendering to provide a broader overview of current methods (\cref{tab:main_results}).
Aiming at high-quality representation of fuzzy geometries, we focus our evaluation on object-centric datasets with prominent fuzzy structures. 
We present results from synthetic benchmark datasets like Shelly \cite{Wang2020SIGGRAPHASIA}, plush objects from real-world tabletop scenes (DTU \cite{jensen2014large}), and additional synthetic custom scenes (our \textit{plushy} and \textit{hairy monkey} from \citet{Sharma2024ECCV}).

Our method, tested on renderings from our real-time Web\-GL renderer, consistently delivers higher image quality than surface-based competitors and renders faster than 3DGS.
We observe that our adaptive shell spacing clusters surfaces around solid structures while maintaining greater separation in volumetric regions (\cref{fig:qualitative}).
Using seven layers offers a good balance between image quality, model size, and rendering speed, as quality tends to degrade with nine meshes under the same number of training iterations. 
This happens as deeper surfaces contribute less to pixel color, reducing gradient magnitudes and slowing optimization.
\cref{fig:results_tradeoffs} illustrates the trade-offs between frame rate on a low-cost smartphone and model size.
Notably, 3DGS fails to meet real-time requirements even when the number of Gaussians is capped during optimization, which leads to a substantial loss in quality.

For comparability, the frame rate of 3DGS is measured using a widely adopted web viewer~\cite{Kellogg2023GaussianSplats3D}. This implementation skips per-frame sorting to prevent slowdowns, using occasional CPU sorting instead. This results in noticeable popping artifacts during rapid camera rotation and suboptimal performance on mobile devices. Our sorting-free method avoids these issues.
While our method falls short of the latest volume-based baselines in image quality, our baked representation provides a favorable balance between quality and speed, rendering much faster on non-specialized hardware.
We refer to our supplementary material for additional results visualizations.



\begin{table}
    \caption{\label{tab:ablations}%
        Ablation studies over intermediate results (5-SDF, implicit representation, before meshes baking) and on final results (5-Mesh, baked, real-time rendering assets).
        See \cref{sec:ablations} for explanations.
        5-Mesh achieves higher quality than the 5-SDF due to its fixed geometry and surface-constrained appearance model; the SDF representation suffers from the stochastic nature of ray sampling, forcing the appearance model to allocate capacity to off-surface elements.
        Results averaged over Shelly \cite{Wang2020SIGGRAPHASIA}.
    }
    \centering
    \resizebox{1.0\columnwidth}{!}{
    \begin{tabular}{l|ccc|ccc}
        \toprule
        & \multicolumn{3}{c|}{5-SDF} & \multicolumn{3}{c}{5-Mesh} \\
        \cmidrule(rl){2-7}
        \textit{Ablation} & PSNR $\uparrow$ & SSIM $\uparrow$ & LPIPS $\downarrow$ & PSNR $\uparrow$ & SSIM $\uparrow$ & LPIPS $\downarrow$ \\ 
        \midrule
        Full  &  32.05 &  0.964 &  0.130 & 34.25 & 0.980 & 0.110  \\
        \midrule
        1) w/o view-dep. $\alpha$  & 31.75 & 0.962 &  0.131  & 32.71 & 0.975 &  0.120  \\
        2) w/o curvature $\mathcal{L}_\text{s}$ & 33.02 & 0.971 &  0.117  & 33.41 &  0.980 &  0.110  \\  
        3) w/o $\alpha_\text{w}$ & 32.11 & 0.966 & 0.130 & 33.96 & 0.980 &  0.113  \\ 
        4) w. const. $\Delta o$ & --- & --- & --- & 30.09 &  0.950 &  0.129  \\
        5) w. outer init. & --- & --- & --- & 30.85 &  0.955 &  0.121  \\
        \bottomrule
    \end{tabular}
    }
\end{table}

\subsection{Ablations}
\label{sec:ablations}
We ablate all crucial aspects of our method and show results on both implicit geometry and fully baked phases in \cref{tab:ablations}. We run our full model:
1) Without view-dependent surface transparency, we observe increased model expressivity as shown by the improved image quality metrics.
2) Without curvature loss during $k$-SDF training ($\lambda_{s} = 0$), surfaces can reconstruct high-frequency details, but the image quality of the baked representation worsens. This happens because the post-baking mesh poorly aligns with the implicit one. Enabling curvature loss pushes $k$-SDF to reconstruct smoother surfaces with meshes that better match their implicit counterparts while keeping the triangle cost low. Moreover, this highlights how our high-capacity appearance model compensates for missing geometric details.
3) Without transparency attenuation (\cref{sec:surfaces_blending}) at grazing angles (no $\alpha_w$), rendering errors increase, especially at object boundaries.
4) With fixed (not trained) support surfaces offsets ($\Delta o$). We observe a significant decline in results, as training offsets is essential for adapting spacing optimally to each scene.
5) Initializing support surfaces outside the main SDF. We observe how this leads geometry to extend beyond the object silhouette. While training views can compensate through learned view-dependent transparency, test views suffer from degraded generalization. In contrast, initializing surfaces inside biases them to be tighter, preventing unwanted expansion. This happens because the reconstructed main surface is typically conservative, forming an outer shell that encloses the scene content, including fuzzy geometry (\cref{fig:teaser}).

\subsection{Limitations}
\label{sec:limitations}
Textured shells \cite{lengyel2000real, lengyel2001real, bears2014practical} exhibit artifacts at grazing angles, especially when test views fall outside training coverage or model capacity is limited. Increasing shell count mitigates this but raises memory and computation costs, particularly during reconstruction.
Artist-designed extruded textured fins \cite{lengyel2001real} can address these artifacts, though learning this component remains challenging and is left for future work.
Our model performs well in densely observed scenes but struggles in sparsely sampled ones. When under-constrained, it tends to explain observations through view-dependency rather than multi-view consistent geometry, leading to poorer generalization in test views.
A solution explored by \citet{wan2023learning} trains a robust model (e.g., NeRF-like) and distills its reconstruction using renderings from randomly generated cameras as the training set.
Handling thin structures remains challenging due to the limitations of the underlying SDF geometry representation.
Advantages on fully solid surfaces are also marginal; see the supplement for further details.

\section{Conclusion}
\label{sec:conclusions}

We presented Volumetric Surfaces, a multi-layer mesh representation for real-time view synthesis of fuzzy objects on a low-power laptops and smartphones. 
Our method renders faster than state-of-the-art volume-based approaches, while being significantly more capable at reproducing fuzzy objects than single-surface methods. 
For future work, we aim to develop a single-stage, end-to-end training procedure that directly generates real-time renderable assets.

\subsection*{Acknowledgments}
Plushy Blender object by \emph{kaizentutorials}.
Stefano Esposito acknowledges travel support from the European Union’s Horizon 2020 research and innovation program under ELISE Grant Agreement No. 951847.




{
    \small
    \bibliographystyle{ieeenat_fullname}
    \bibliography{bibliography_long,bibliography,bibliography_custom}
}


\clearpage

\beginsupplement

\setcounter{page}{1}
\maketitlesupplementary


\noindent
In this supplementary material, we provide additional architectural and technical details (\cref{sec:technical}), further visualizations (\cref{sec:additional_visualizations}), comprehensive per-scene results (\cref{sec:per-scene-results}), and an in-depth analysis of performance on fully solid geometries (\cref{sec:solid-scenes}).

\section{Additional Technical Information}
\label{sec:technical}

\boldparagraph{$\beta$ scheduling details}
During training, the $\beta$ parameter is controlled by the scheduling of $v$ as $\beta = e^{10v}$. 
During the main surface training phase, $v$ linearly transitions from $v_1 = 0.3$ to $v_2 = 0.7$. 
During the training of $k$-SDF, it further progresses from $v_2  = 0.7$ to $v_3 = 1.0$.
At $\beta_2$, the logistic distribution standard deviation is approximately $0.001$. We use this value to initialize offsets as constants ($\Delta o$). By the end of implicit surface training ($\beta_3$), it decreases to $0.00008$, resulting in fully peaked densities.

\boldparagraph{$k$-SDF Architecture}
We encode 3D points using the trainable positional encoding from \citet{Rosu2023CVPR}, followed by a small MLP with three layers of 32 features each. Hidden layers employ \texttt{GELU} activations, while the final layer uses a linear activation to output the signed distance $d$ (our main SDF) and a geometric feature vector $\mathbf{z}$.
We predict relative offsets using tiny MLP heads (a single layer with 32 units) with independent parameters, taking only $\mathbf{z}$ as input. This ensures that model complexity scales with the number of surfaces.
To enforce the sign of the predicted offset, we apply a \texttt{softplus} activation multiplied by the desired sign.
Finally, we compute the final ordered offsets by performing a cumulative sum over the predicted relative offsets, separately for negative and positive values.

\boldparagraph{Volumetric Appearance Architectures}
We model RGB and transparency as two networks with identical architectures, differing only in their output dimensions (3 for RGB and 1 for transparency). Both models encode 3D points using the trainable positional encoding from \citet{Rosu2023CVPR}, followed by an MLP with three layers of 128, 128, and 64 features, respectively. Its input consists of the encoded position, a spherical harmonics encoding (with degree 3) of the view direction $\mathbf{v}$, the normal vector of the rendered SDF $\mathbf{n}$ and the geometric feature vector $\mathbf{z}$ predicted by $k$-SDF. Normals are computed as the normalized gradients of the SDFs; gradients are computed with finite differences, $\epsilon = 10^{-4}$. Hidden layers use \texttt{GELU} activations, while the final layer applies a \texttt{Sigmoid} activation to produce outputs in the range $[0, 1]$.
 
\boldparagraph{Neural Textures Architecture}
During the mesh texturing phase, we use separate neural texture models for RGB and transparency per mesh. We encode 2D UV coordinates using the trainable positional encoding from \citet{Mueller2022SIGGRAPH}, followed by a small MLP with two layers of 64 features each. Hidden layers use \texttt{ReLU} activations, while the final layer applies a linear activation to output per-channel spherical harmonics (SH) coefficients of degree 3, which are then decoded with view direction $\mathbf{v}$.

\boldparagraph{Points sampling}
%
During volumetric rendering the number of uniformly sampled points per ray in the foreground area of the scene is 64. On top of these, 32 points are added with importance sampling. Additionally, if a scene is unbounded (e.g. DTU), we sample 32 additional points in contracted space \cite{Barron2022CVPR}.
%
Rays batch size is defined w.r.t. a target number of sample points which is up to a maximum of $512 \times 64 \times 32$ points.

\boldparagraph{Data handling}
We use \texttt{MVDatasets}~\cite{Esposito2024MVDatasets} to load datasets, manage training loop pixel iterators, and perform ray casting.


\begin{figure*}
    \centering
    \begin{tabular}{c}
        \includegraphics[width=0.99\textwidth]{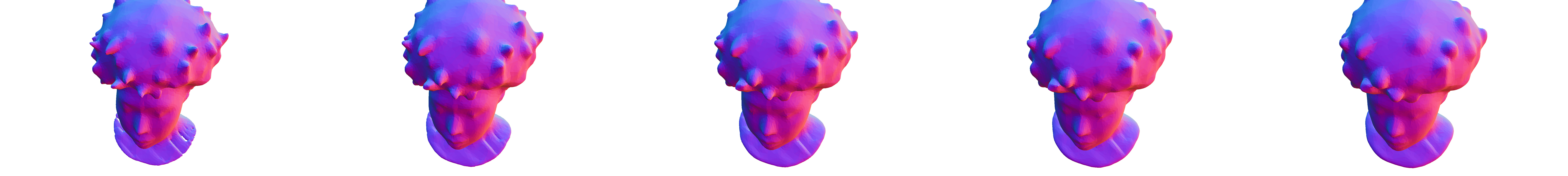} \\
        a) Surface normals. \vspace{0.5em} \\
        \includegraphics[width=0.99\textwidth]{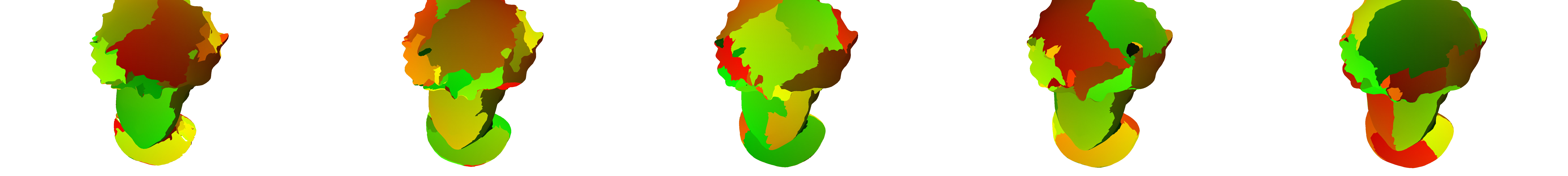} \\
        b) Surface UVs. \vspace{0.5em} \\
        \includegraphics[width=0.99\textwidth]{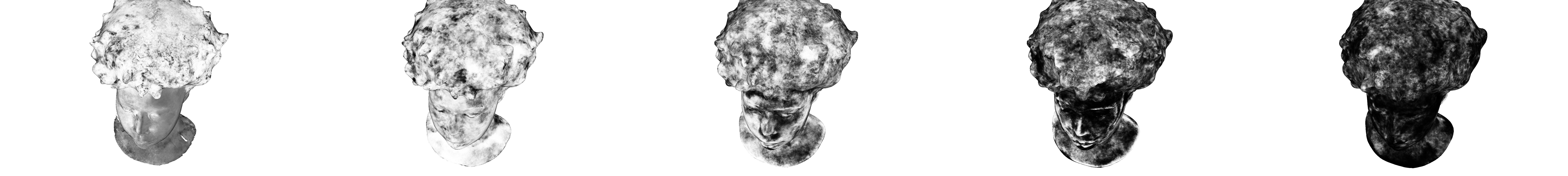} \\
        c) Surface opacity. \vspace{0.5em} \\
        \includegraphics[width=0.99\textwidth]{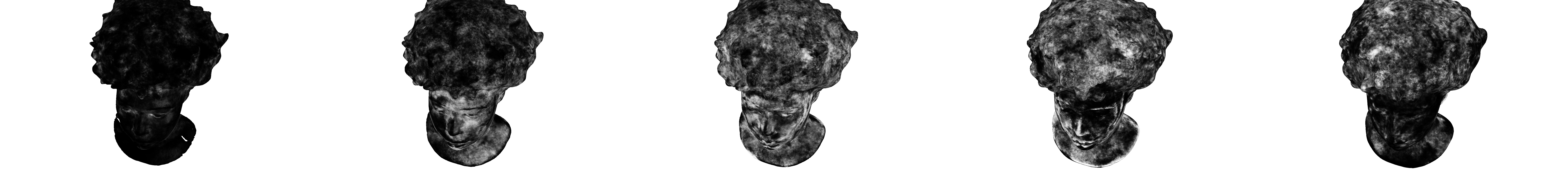} \\
        d) Blending weights (contributions). \vspace{0.5em} \\
        \includegraphics[width=0.99\textwidth]{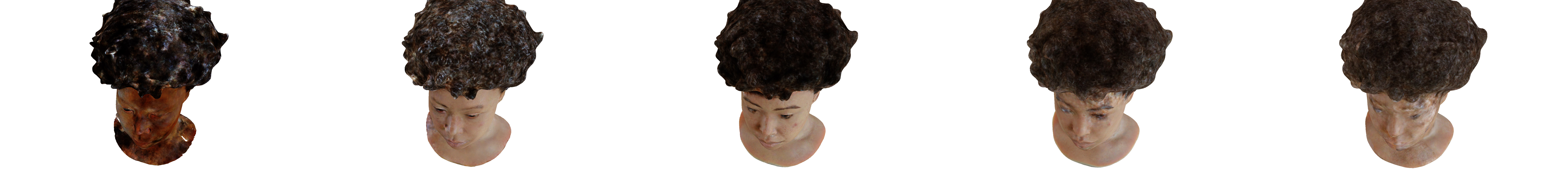} \\
        e) Surface colors (RGB). \vspace{0.5em} \\
    \end{tabular}
    \caption{
    Visualization of render buffers from our 5-Mesh model. Layers order: left to right is inner to outer.
    Individual layer color and alpha buffers are blended as described in \cref{sec:surfaces_blending}.
    Results on the \emph{khady} scene from Shelly \cite{Wang2020SIGGRAPHASIA}. 
    }
    \label{fig:buffers_2}
\end{figure*}

\begin{figure}
    \centering
    \begin{tabular}{cc} 
        \includegraphics[width=0.45\columnwidth]{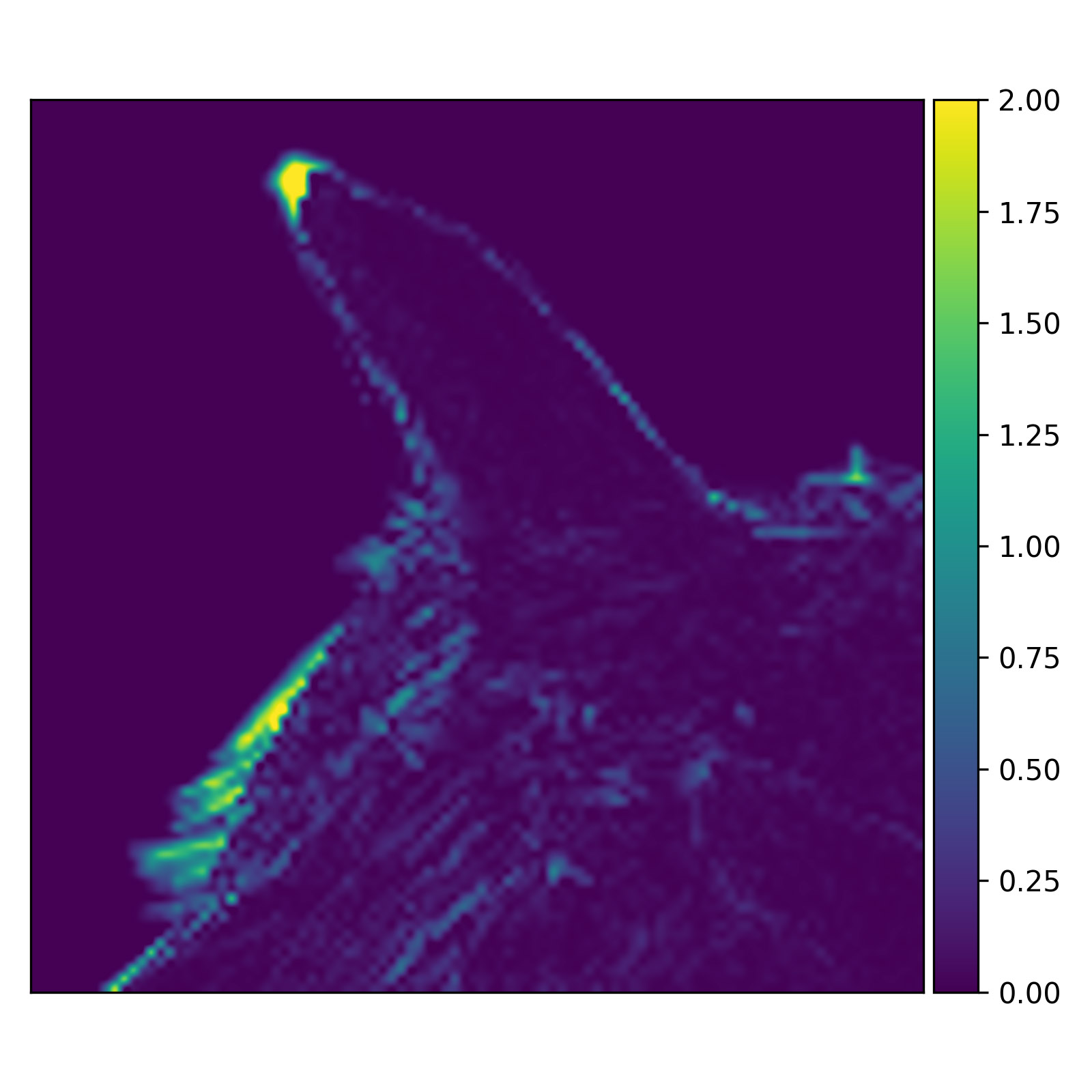} & 
        \includegraphics[width=0.45\columnwidth]{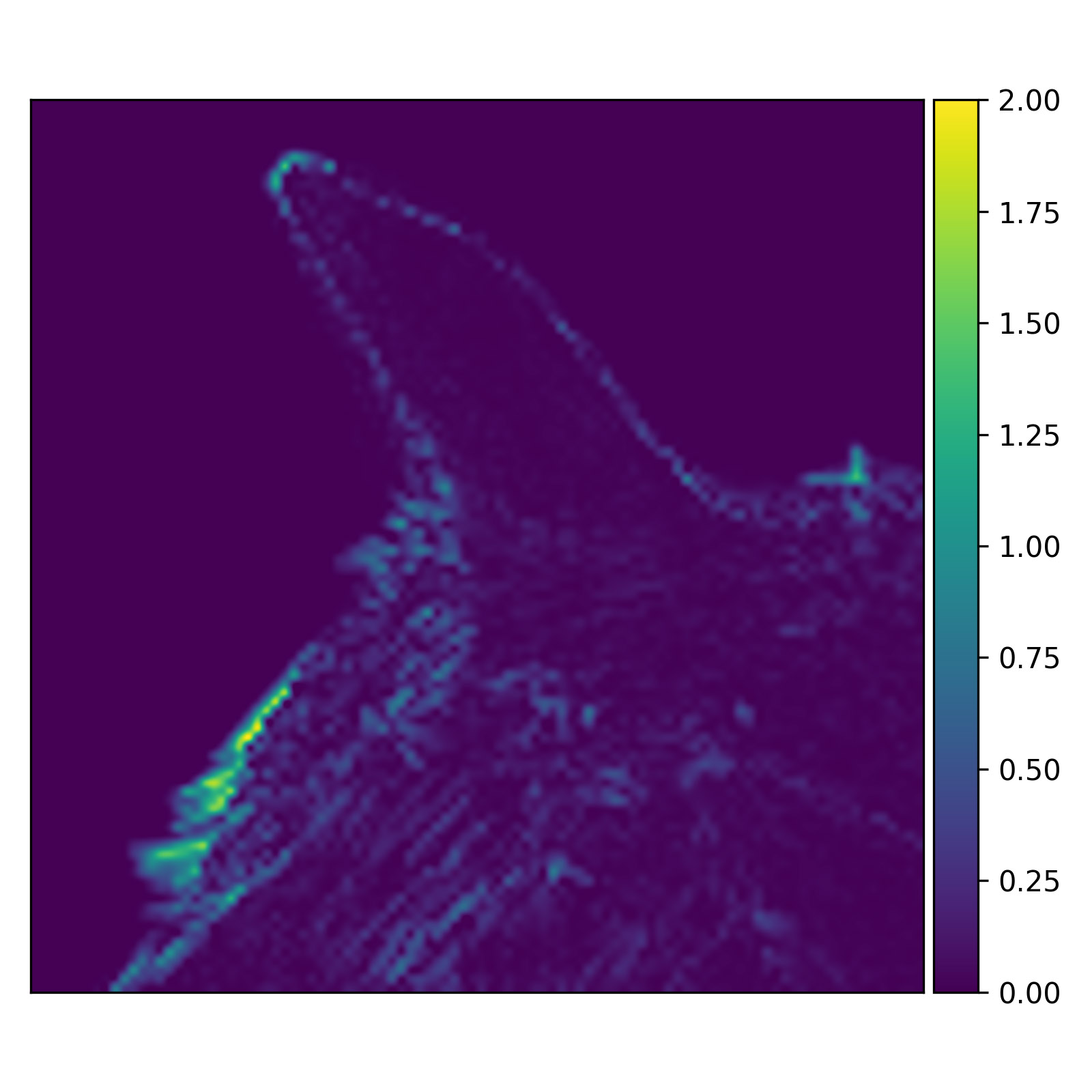} \\
        $a$) without $\alpha_\text{w}$ & $b$) with $\alpha_\text{w}$  
    \end{tabular}
    \caption{
    ($a$) Rendering error crop (averaged over color channels) without and ($b$) with transparency-decay, resulting in a 2.13\,dB PSNR gain.
    Scene from the Shelly \cite{Wang2020SIGGRAPHASIA}.
    }
    \label{fig:alpha_attenuation}
\end{figure}

\begin{figure*}
    \centering
    \begin{tabular}{c}
        \includegraphics[width=0.99\textwidth]{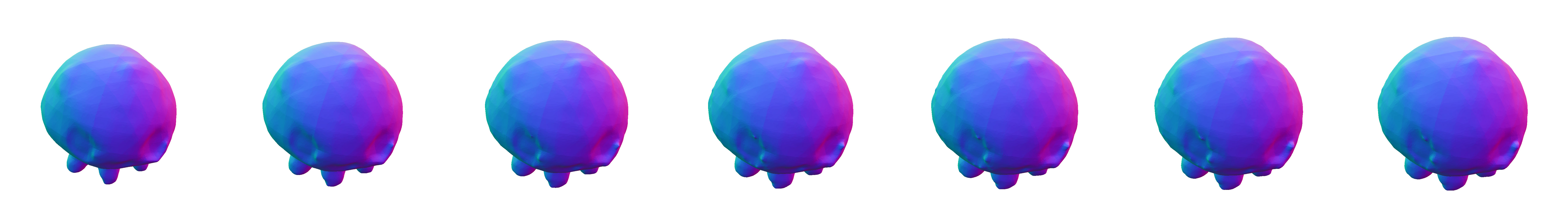} \\
        a) Surface normals. \vspace{0.5em} \\
        \includegraphics[width=0.99\textwidth]{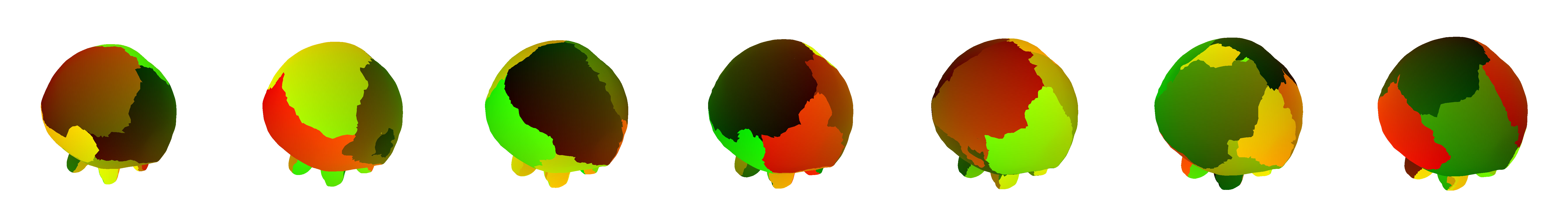} \\
        b) Surface UVs. \vspace{0.5em} \\
        \includegraphics[width=0.99\textwidth]{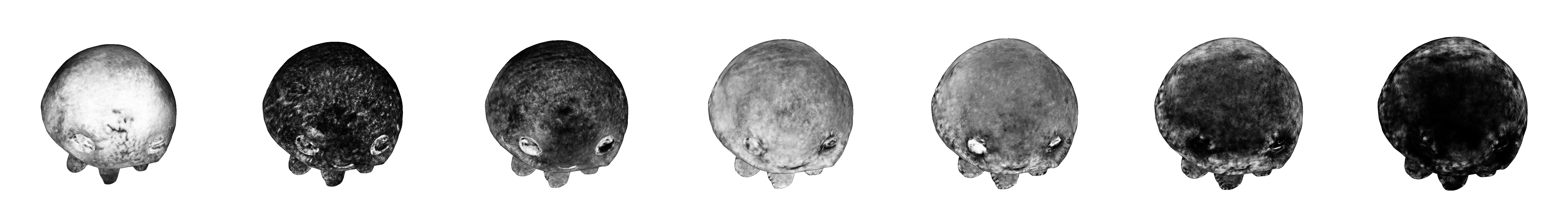} \\
        c) Surface opacity. \vspace{0.5em} \\
        \includegraphics[width=0.99\textwidth]{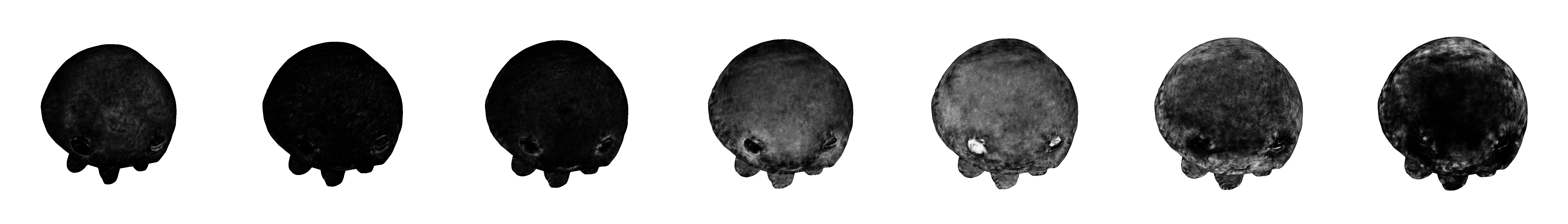} \\
        d) Blending weights (contributions). \vspace{0.5em} \\
        \includegraphics[width=0.99\textwidth]{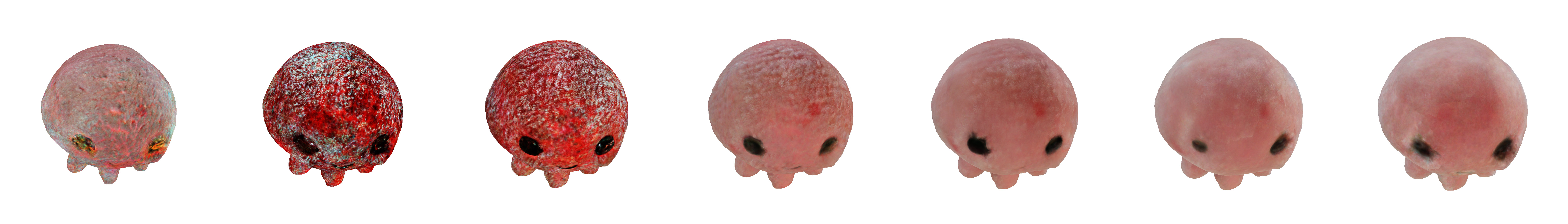} \\
        e) Surface colors (RGB). \vspace{0.5em} \\
    \end{tabular}
    \caption{
    Visualization of render buffers from our 7-Mesh model. Layers order: left to right is inner to outer.
    Individual layer colors and alpha buffers are blended as described in \cref{sec:surfaces_blending}.
    Results our custom \emph{plushy} scene. 
    }
    \label{fig:buffers_1}
\end{figure*}


\section{Additional Visualizations}
\label{sec:additional_visualizations}

\begin{figure*}
    \centering
    \begin{tabular}{cc} 
        \includegraphics[width=0.33\textwidth]{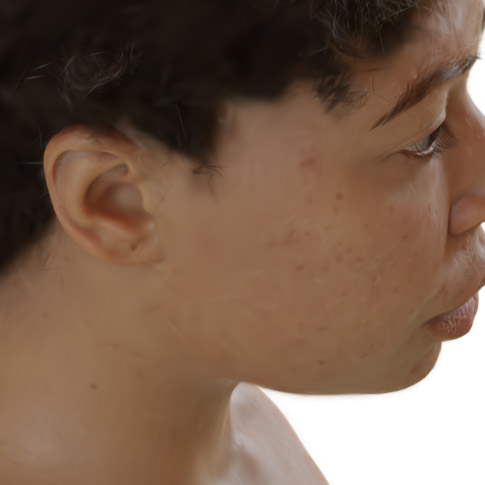} & 
        \includegraphics[width=0.33\textwidth]{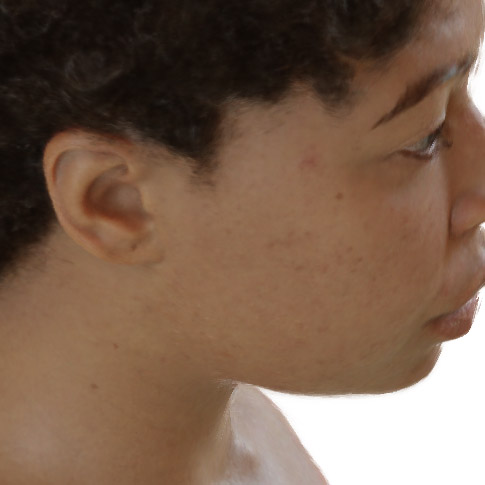} \\
        $a$) 3DGS & $b$) 7-Mesh (ours) \\[1em]
        \includegraphics[width=0.33\textwidth]{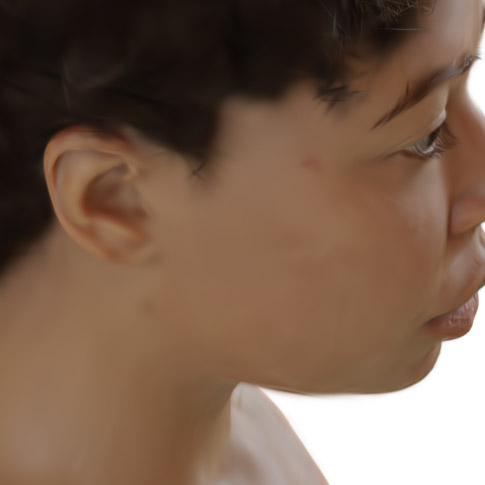} &
        \includegraphics[width=0.33\textwidth]{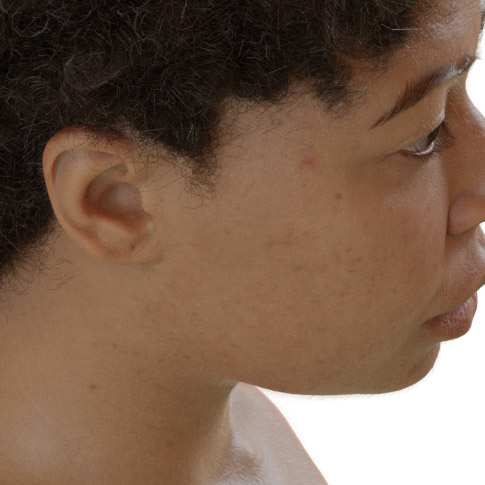} \\
        $c$) 3DGS-75K & $d$) Ground truth 
    \end{tabular}
    \caption{
    3DGS~\cite{Kerbl2023SIGGRAPH} demonstrates superior performance in modeling thin structures but is significantly less effective in representing large, textured areas. Our method renders faster than 3DGS-75K on mobile devices.
    Results on the \emph{khady} scene from Shelly \cite{Wang2020SIGGRAPHASIA}. 
    Quantitative results are in \cref{tab:performance}.
    }
    \label{fig:3dgs_qualitative}
\end{figure*}

We provide additional qualitative comparisons on our evaluation scenes of the DTU~\cite{jensen2014large} dataset.
Additionally, we provide visualizations of per-surface rendering before alpha blending (\cref{fig:buffers_2} and \cref{fig:buffers_1}) to illustrate how each layer, based on its position and opacity, contributes with its view-dependent appearance model to the final image.
Finally, we visualize results from \cref{tab:performance}. 
\cref{fig:3dgs_qualitative} presents a qualitative comparison between a render of our 7-Mesh model, 3DGS~\cite{Kerbl2023SIGGRAPH}, and 3DGS-75K.
\cref{fig:mobilenerf_qualitative} compares our 5-Mesh model to MobileNeRF~\cite{chen2023mobilenerf}.

\subsection{Transparency Attenuation}
\label{sec:transp_atten_supp}

We introduced transparency attenuation in \cref{sec:surfaces_blending} to reduce visual artifacts at object boundaries.
\cref{fig:alpha_attenuation}, cropped from our ablation experiments (\cref{sec:ablations}), highlights its significance in our method.


\section{Per-scene Results}
\label{sec:per-scene-results}

\cref{tab:scenes_ours_baselines}, \cref{tab:scenes_ours_1} and \cref{tab:scenes_ours_2} present the per-scene values that are averaged in \cref{tab:main_results}.


\section{Fully Solid Scenes}
\label{sec:solid-scenes}

\begin{figure*}
    \centering
    \begin{tabular}{ccc} 
        \includegraphics[width=0.32\textwidth]{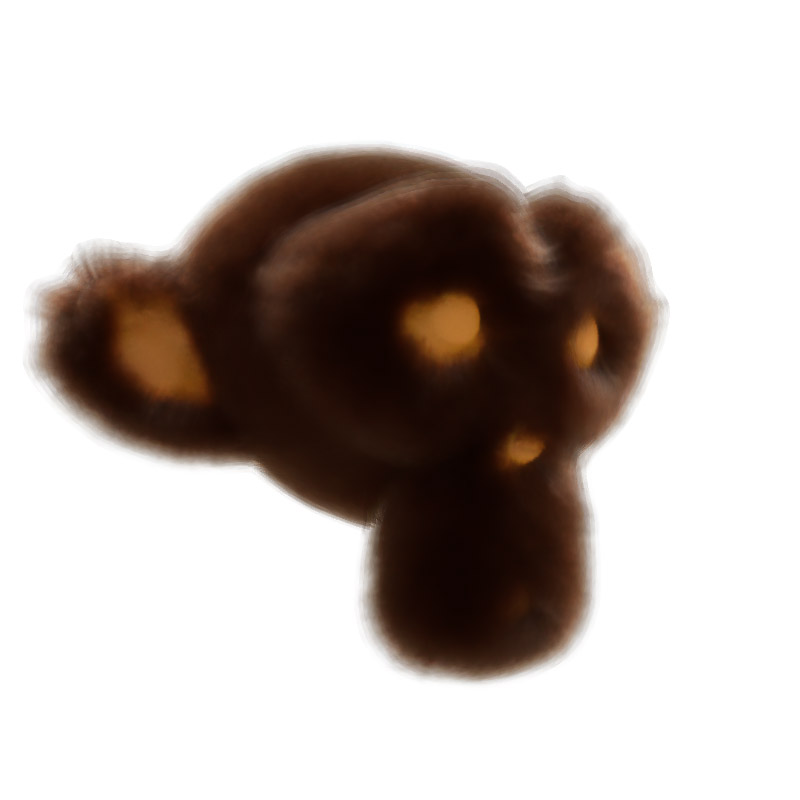} &
        \includegraphics[width=0.32\textwidth]{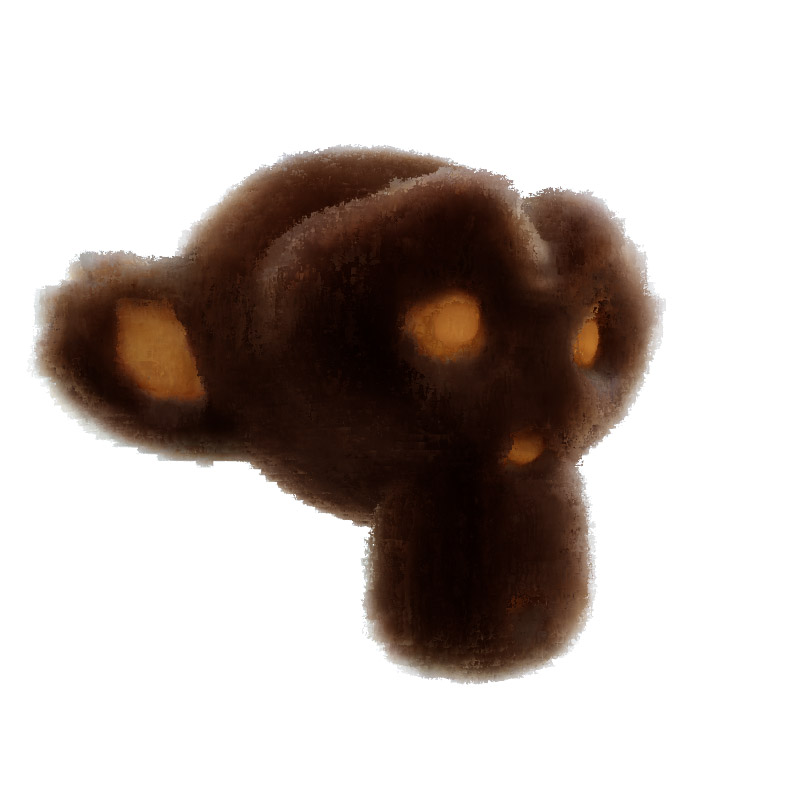} &
        \includegraphics[width=0.32\textwidth]{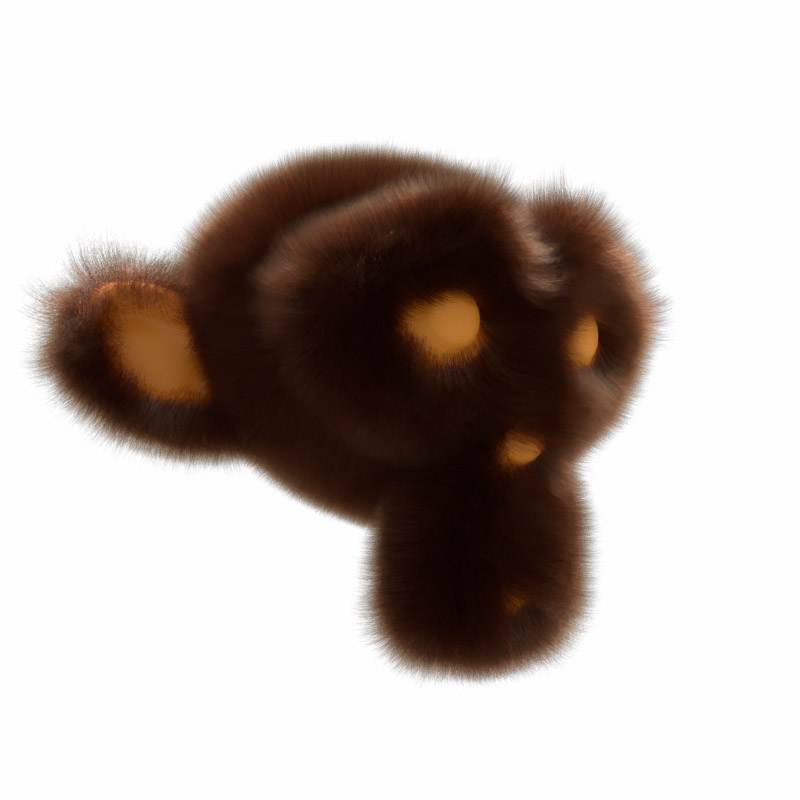} \\ [0.5em] 
        \includegraphics[width=0.32\textwidth]{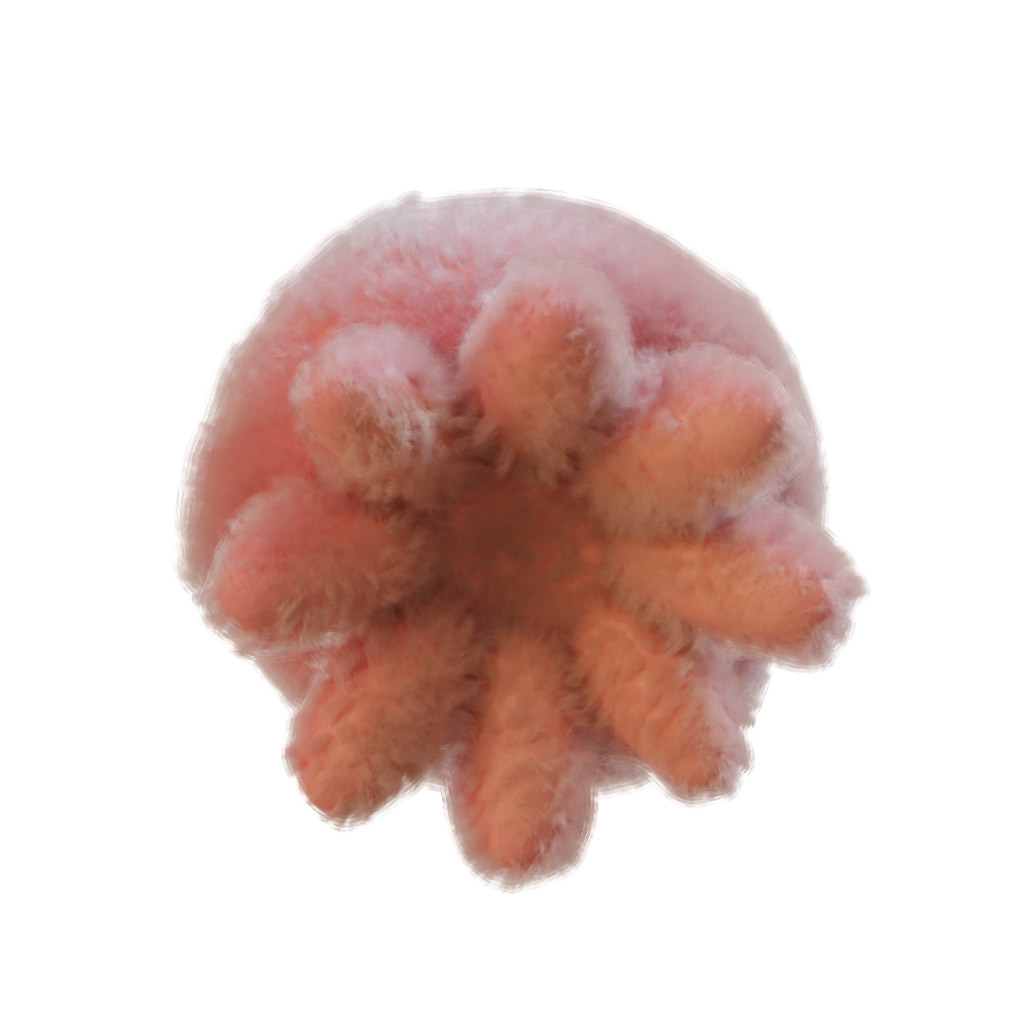} &
        \includegraphics[width=0.32\textwidth]{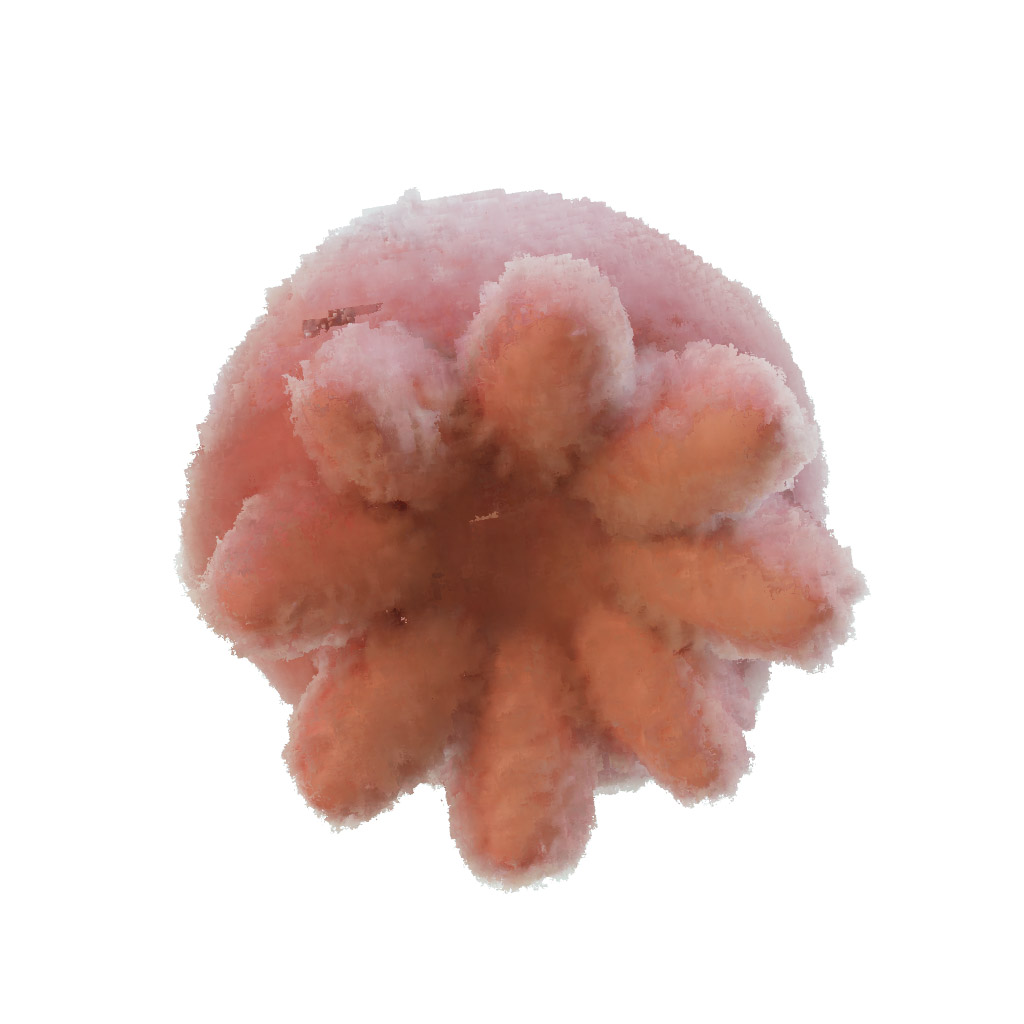} &
        \includegraphics[width=0.32\textwidth]{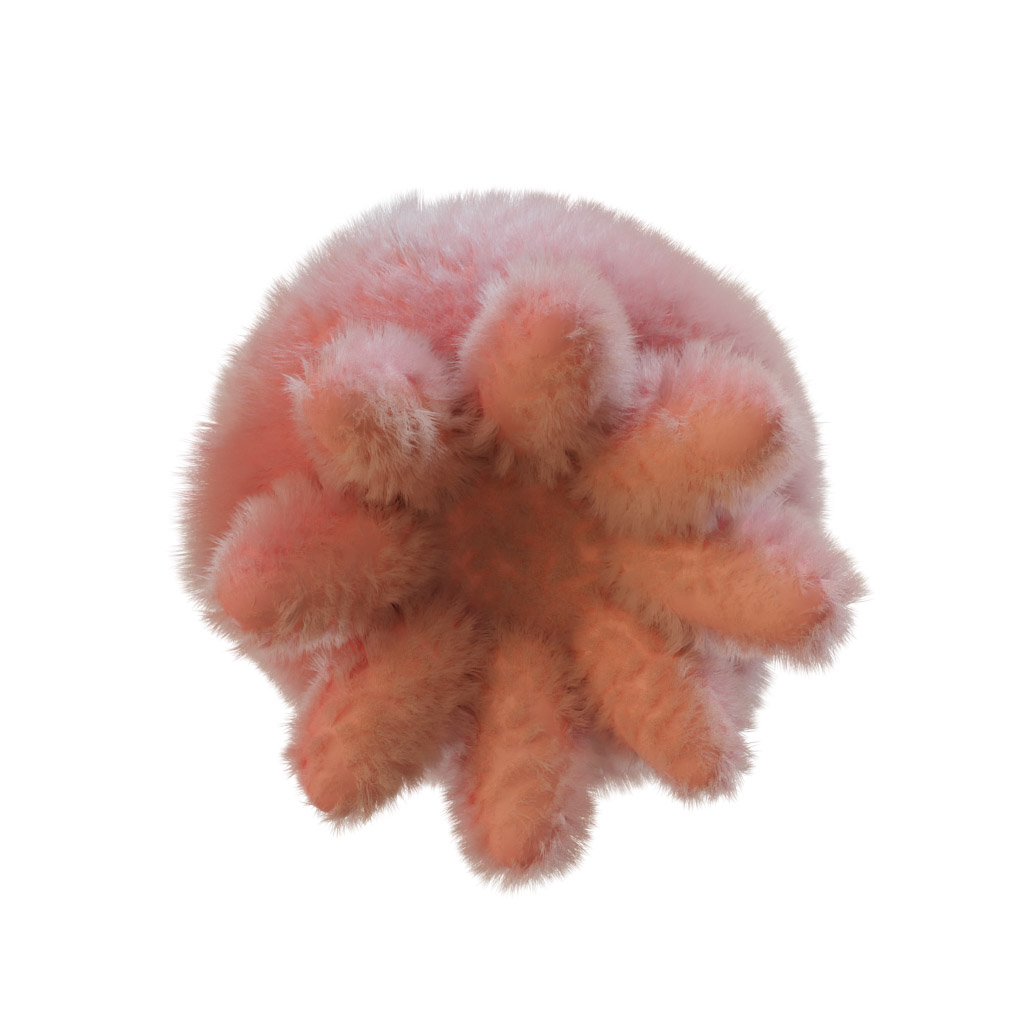} \\ [0.5em] 
        \includegraphics[width=0.32\textwidth]{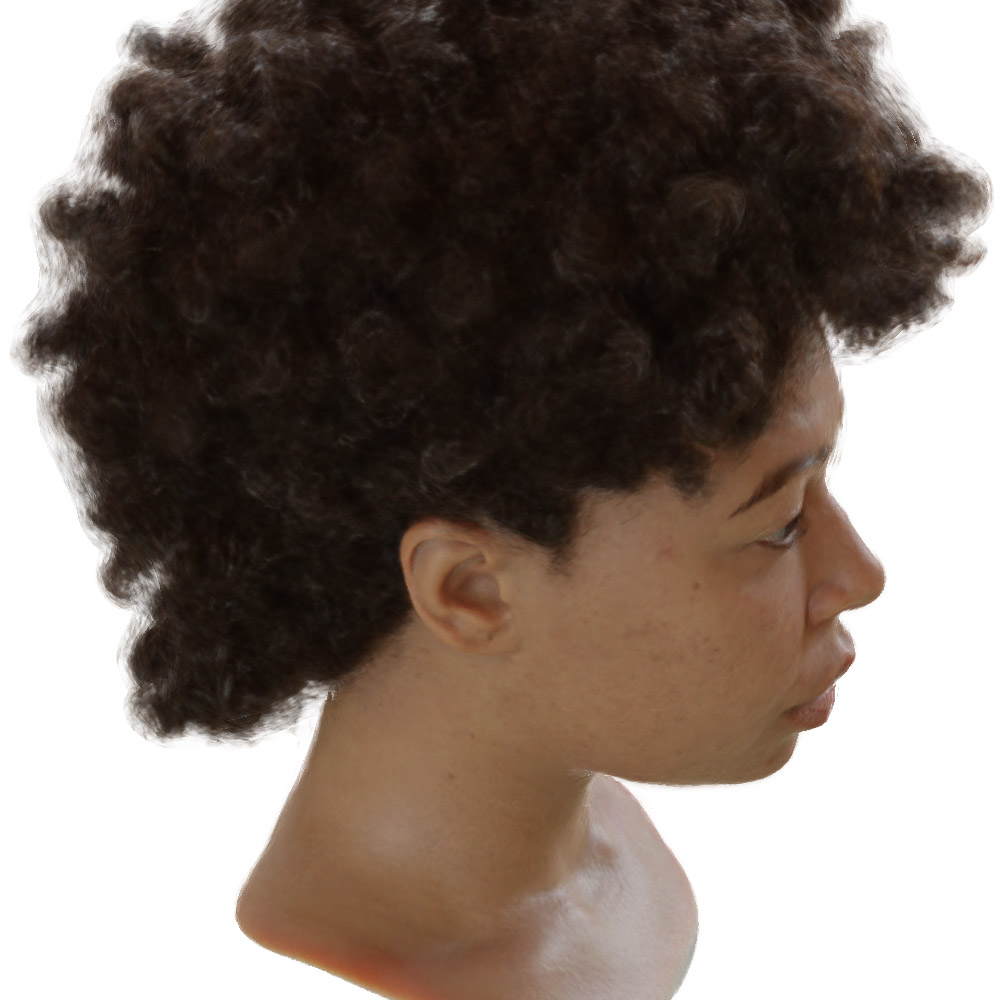} &
        \includegraphics[width=0.32\textwidth]{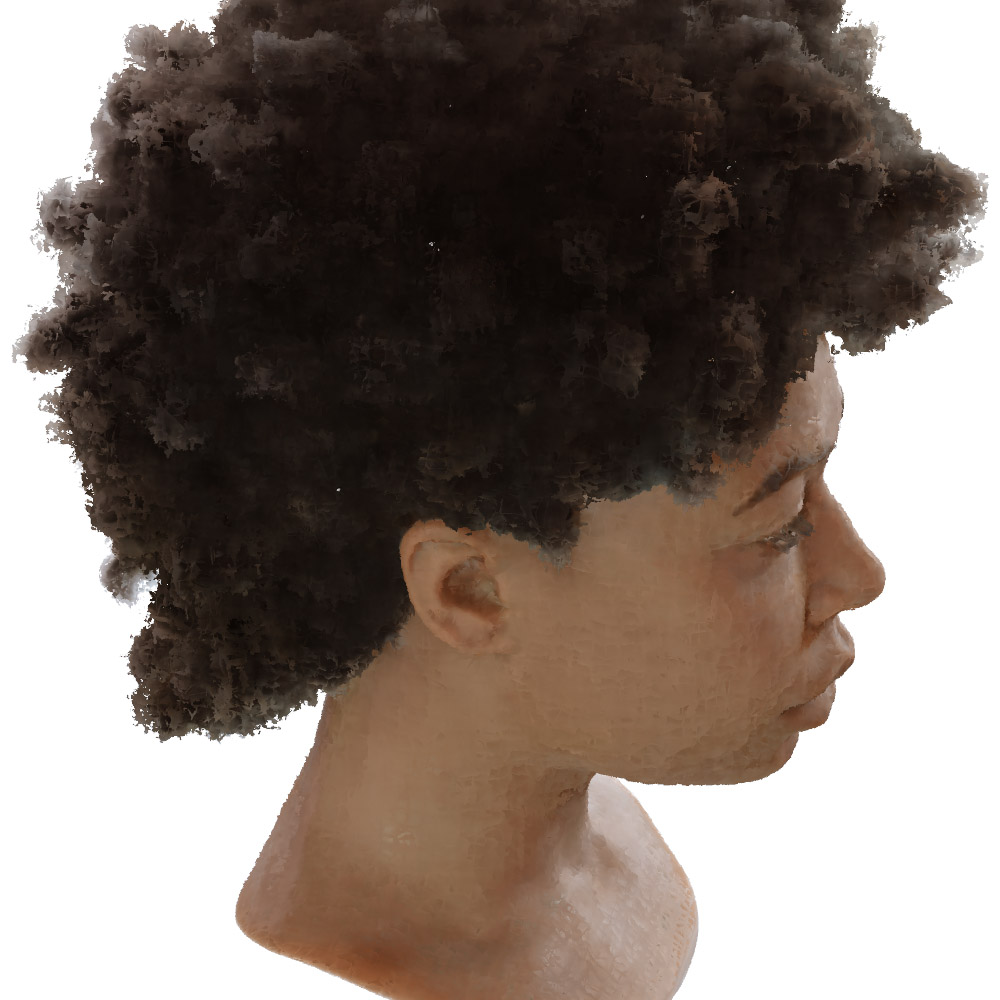} &
        \includegraphics[width=0.32\textwidth]{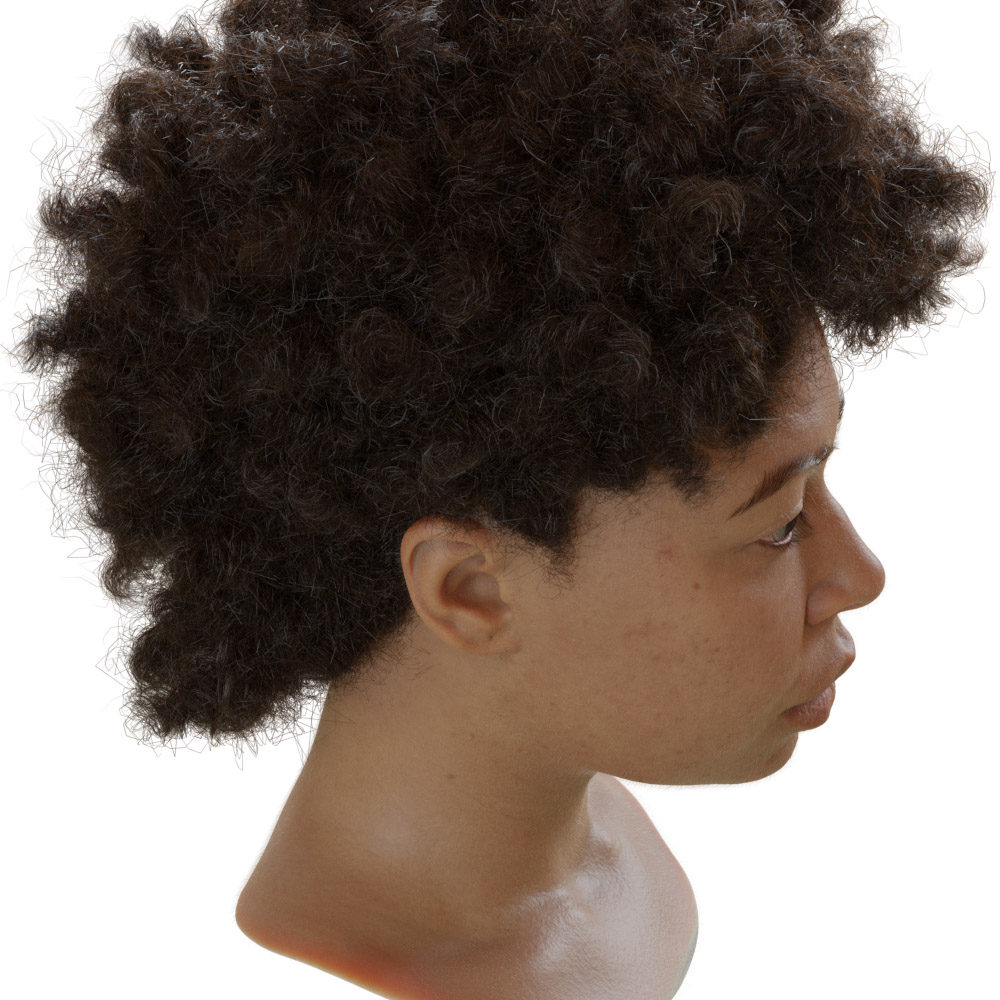} \\
        $a$) 5-Mesh (ours) & $b$) MobileNeRF & $c$) Ground truth
    \end{tabular}
    \caption{
    Our method surpasses MobileNeRF~\cite{chen2023mobilenerf} in modeling volumetric hair while also achieving superior performance on flat surfaces.
    Results on \emph{hairy monkey} from QuadFields~\cite{Sharma2024ECCV}, our custom \emph{plushy} scene and \emph{khady} from Shelly \cite{Wang2020SIGGRAPHASIA}.
    Quantitative results are in \cref{tab:performance}.
    }
    \label{fig:mobilenerf_qualitative}
\vspace{0.5em}
    \centering
    \includegraphics[width=0.99\textwidth]{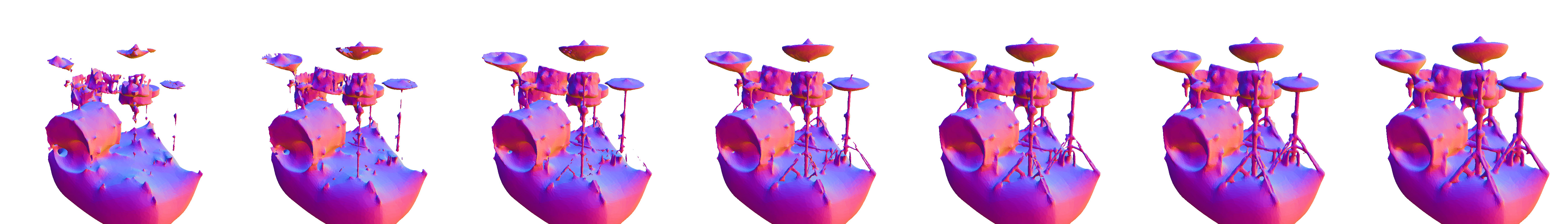}
    \caption{\label{fig:blender-normals}%
    Visualization of surface normals from our 7-Mesh model. Scene from NeRF-Synthetic~\cite{Mildenhall2020ECCV}.}
\end{figure*}

\begin{figure*}
    \centering
    \resizebox{\textwidth}{!}{
    \begin{tabular}{cccc} 
        \includegraphics[width=0.25\textwidth]{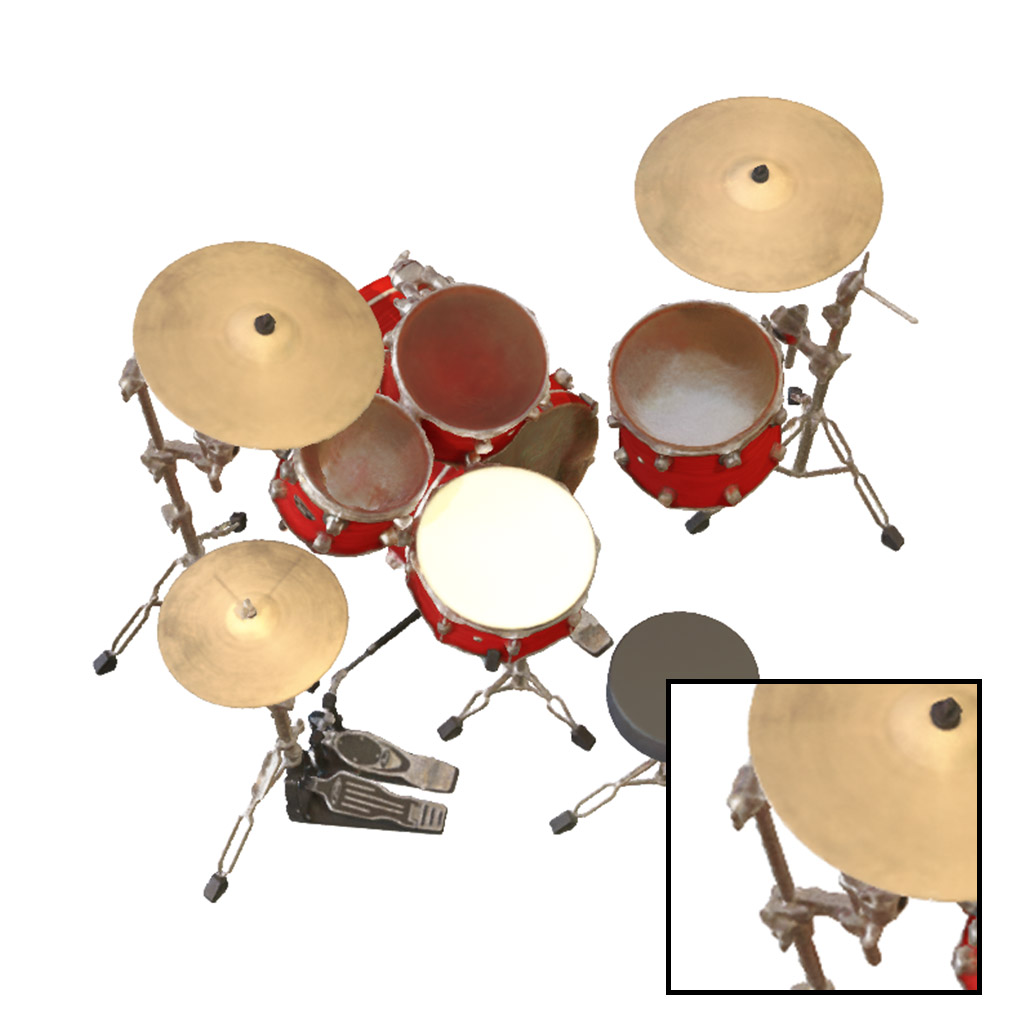} & 
        \includegraphics[width=0.25\textwidth]{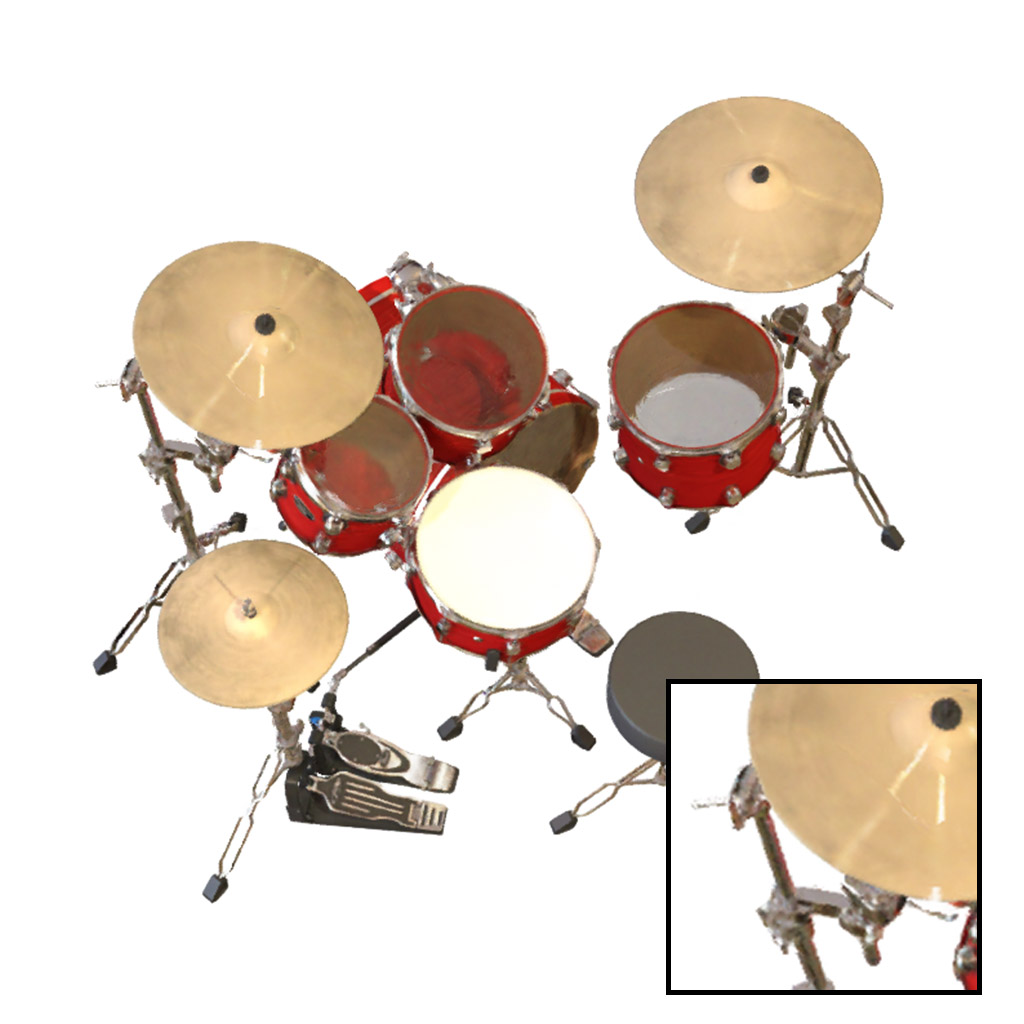} & 
        \includegraphics[width=0.25\textwidth]{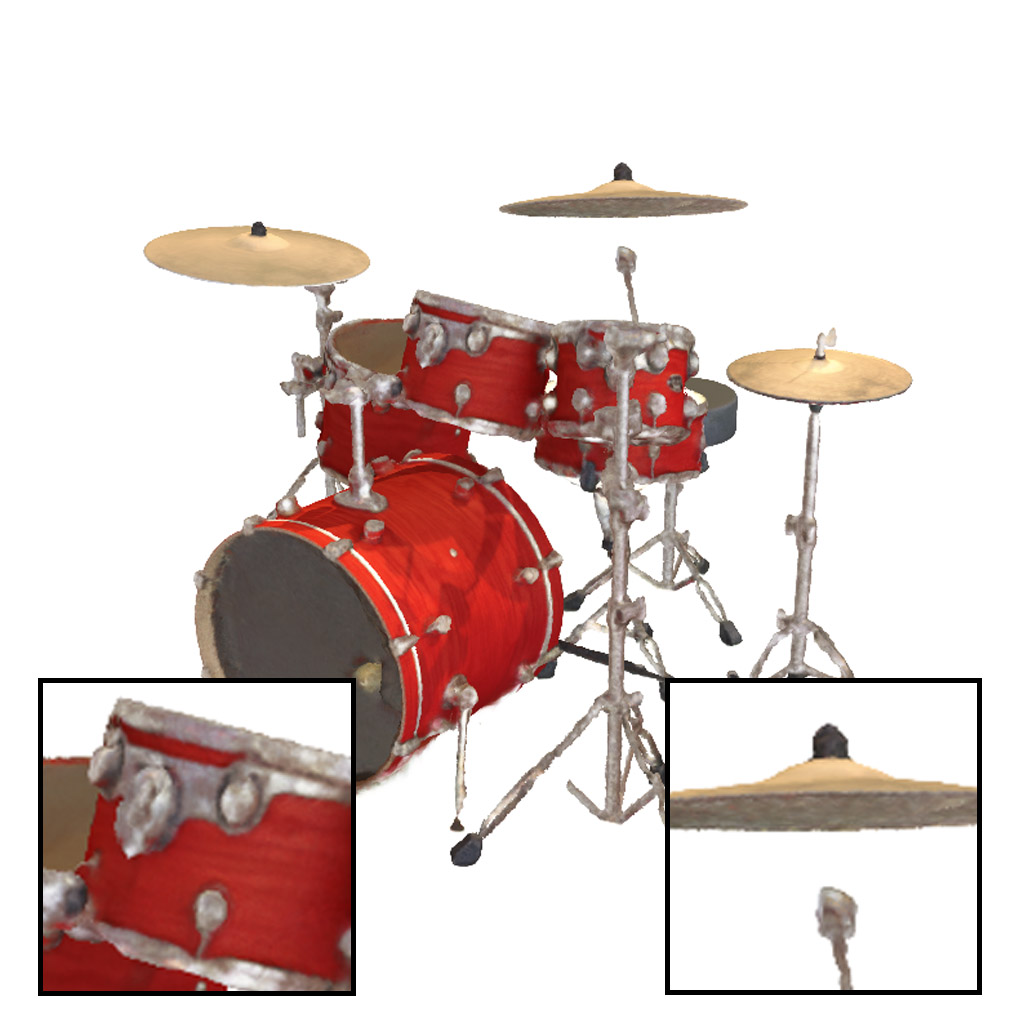} & 
        \includegraphics[width=0.25\textwidth]{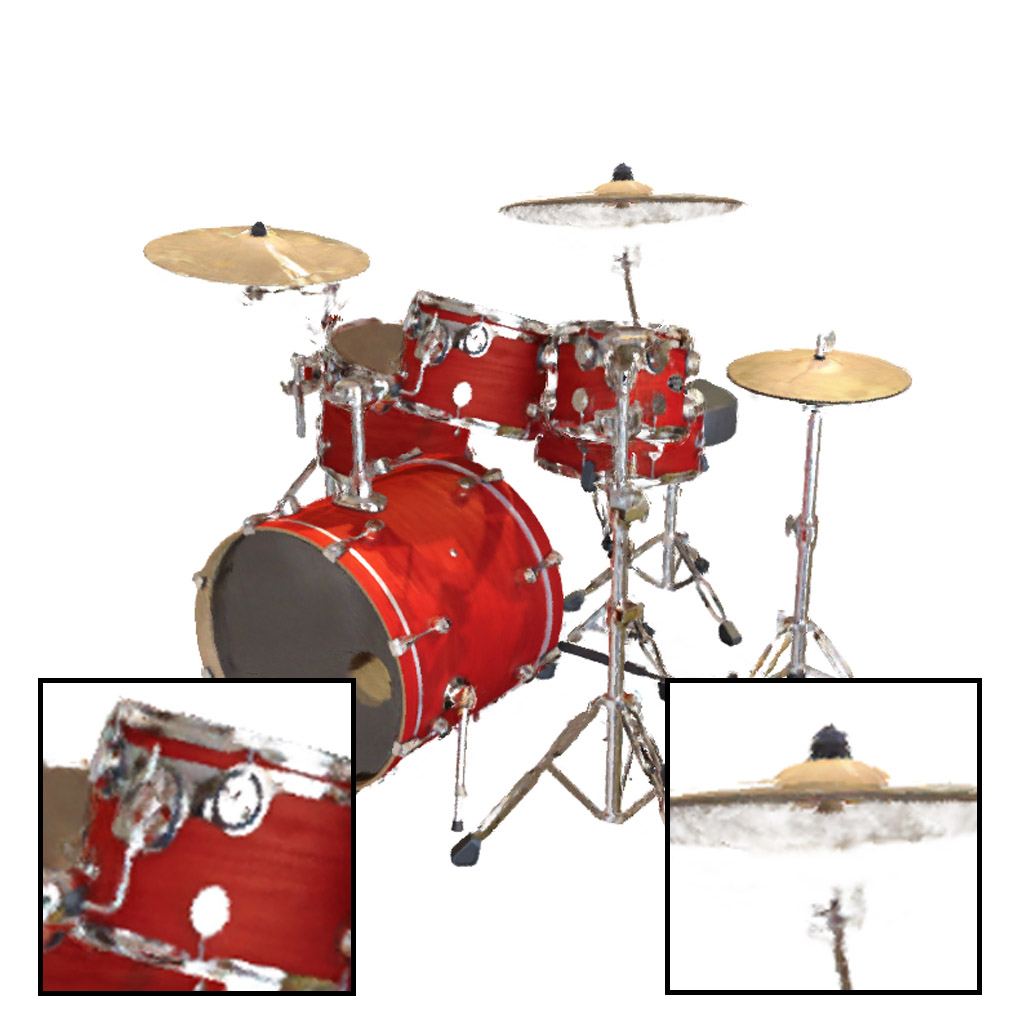} \\
        $a$) PermutoSDF & $b$) 7-Mesh (ours) & $c$) PermutoSDF & $d$) 7-Mesh (ours) \\
    \end{tabular}
    } 
    \caption{\label{fig:blender_qualitative}%
    Qualitative comparison of our method with PermutoSDF \cite{Rosu2023CVPR}. Renderings ($a$) and ($b$) are from a training view.
    Renderings ($c$) and ($d$) are from a test view.
    Scene from NeRF-Synthetic \cite{Mildenhall2020ECCV}.
    }
\end{figure*}

Although not our targeted use case, we tested our method on the fully solid scenes of the NeRF-Synthetic dataset~\cite{Mildenhall2020ECCV}, which lacks fuzzy objects. As noted in \cref{sec:limitations}, our advantages in these scenes are marginal. While we outperform PermutoSDF~\cite{Rosu2023CVPR}, our quality remains behind other baselines.
We model fuzzy surfaces by optimizing sample distribution rather than reconstructing high-frequency geometric details, as spatial sampling is key to accurately capturing these effects.
By favoring smoother surfaces, our method tends to reconstruct overly simplified geometry in under-observed areas (\cref{fig:blender-normals}).
Fully solid scenes can be optimally modeled as a single surface.
However, SDF-based methods struggle in handling thin structures, as optimization often fails to reconstruct them reliably (e.g., BakedSDF~\cite{Yariv2023ARXIV}, BOG~\cite{Reiser2024SIGGRAPH}).
Our surface smoothing, combined with view-dependent transparency, often leads to thin structures being reconstructed as view-dependent effects (\cref{fig:blender_qualitative}).
As a result, our model tends to overfit training views, leading to a larger quality gap between training and test views (\cref{tab:quantitative_blender}).



\begin{figure*}
    \centering
    \resizebox{\textwidth}{!}{
    \begin{tabular}{cccc} 
        
        \includegraphics[width=0.27\textwidth]{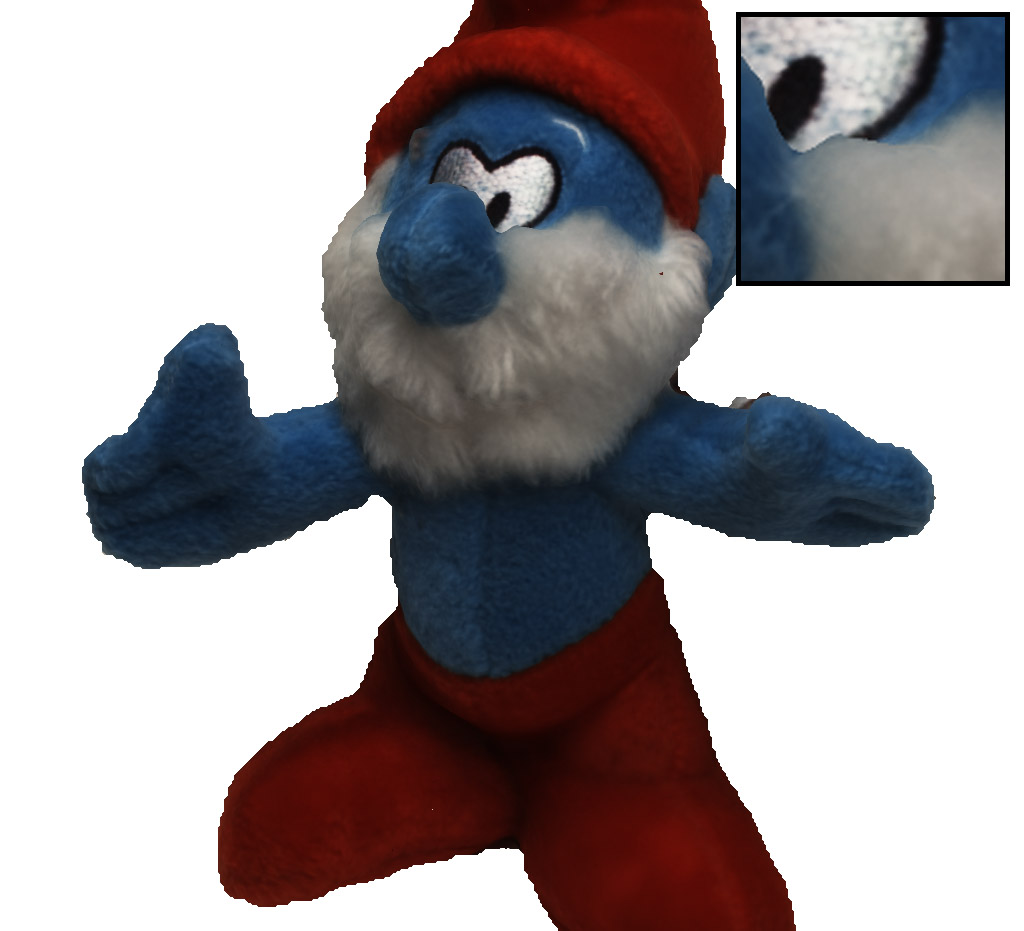} & 
        \includegraphics[width=0.27\textwidth]{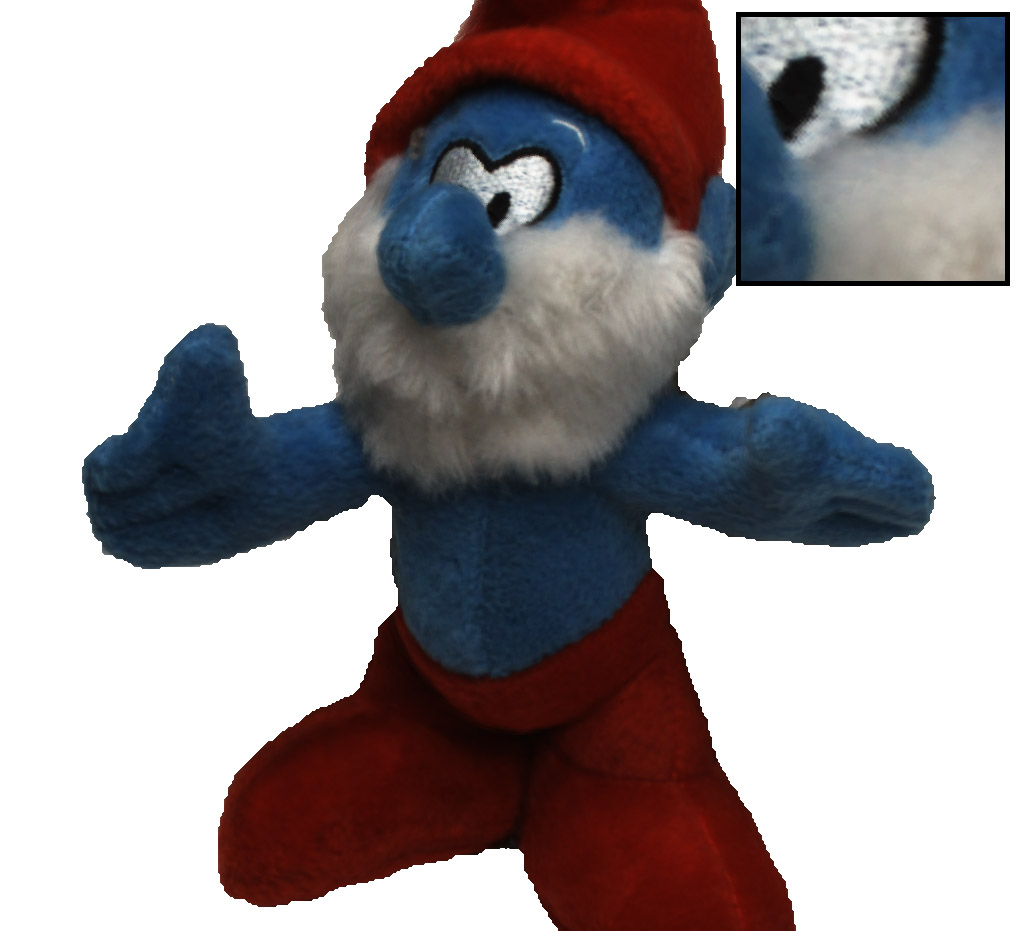}
        &
        \includegraphics[width=0.27\textwidth]{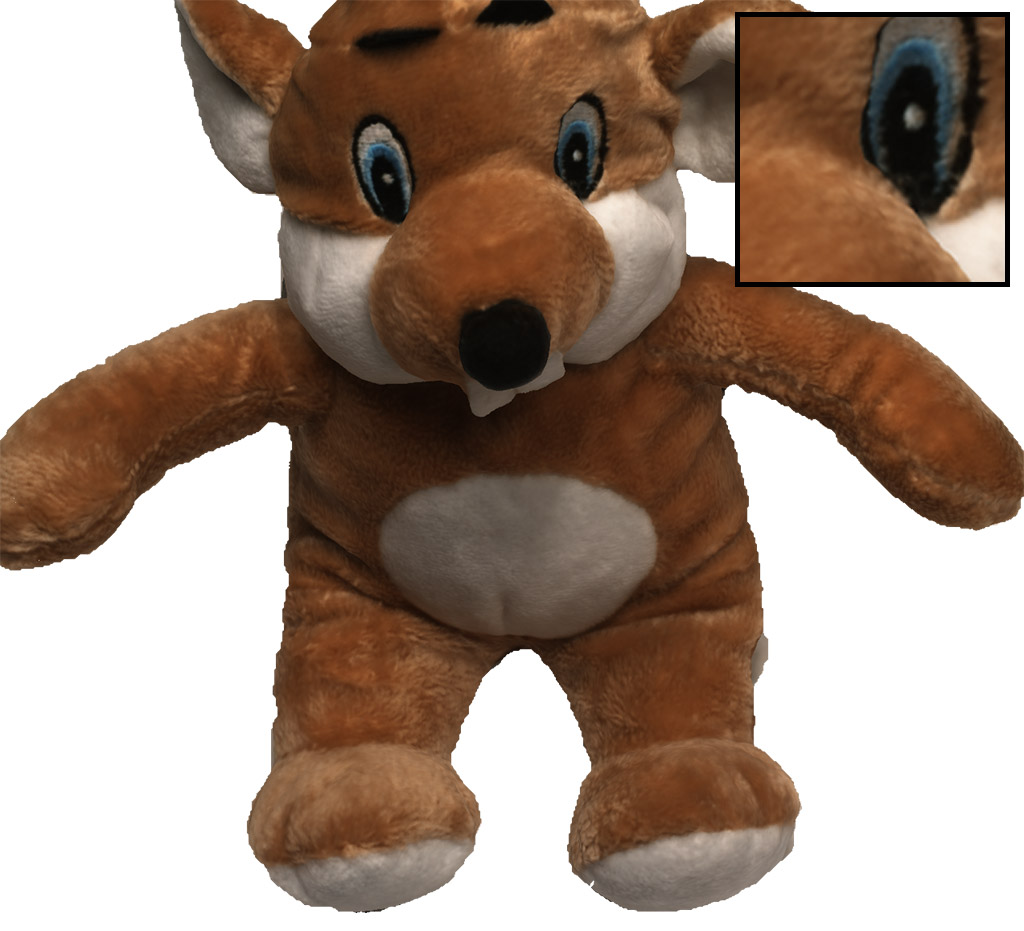} & 
        \includegraphics[width=0.27\textwidth]{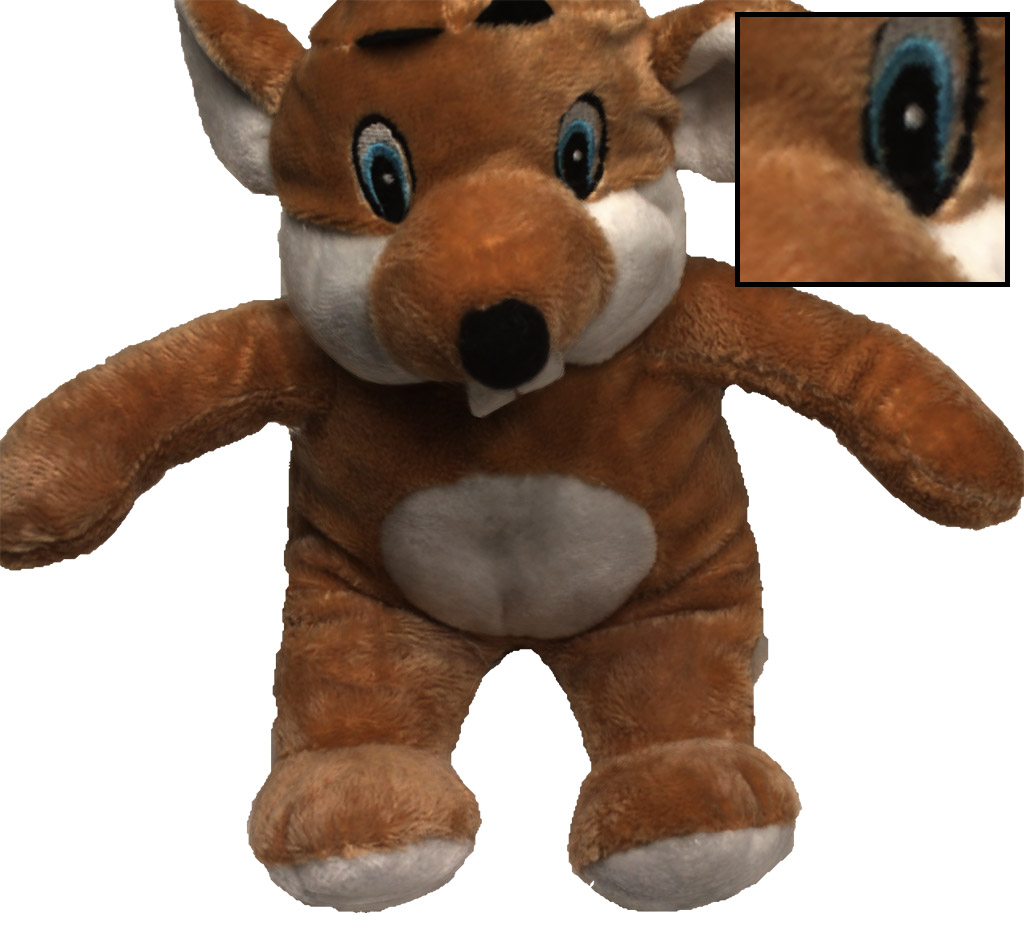} \\ 
         PermutoSDF & 7-Mesh (ours) & PermutoSDF & 7-Mesh (ours) \\[1em]
        \includegraphics[width=0.27\textwidth]{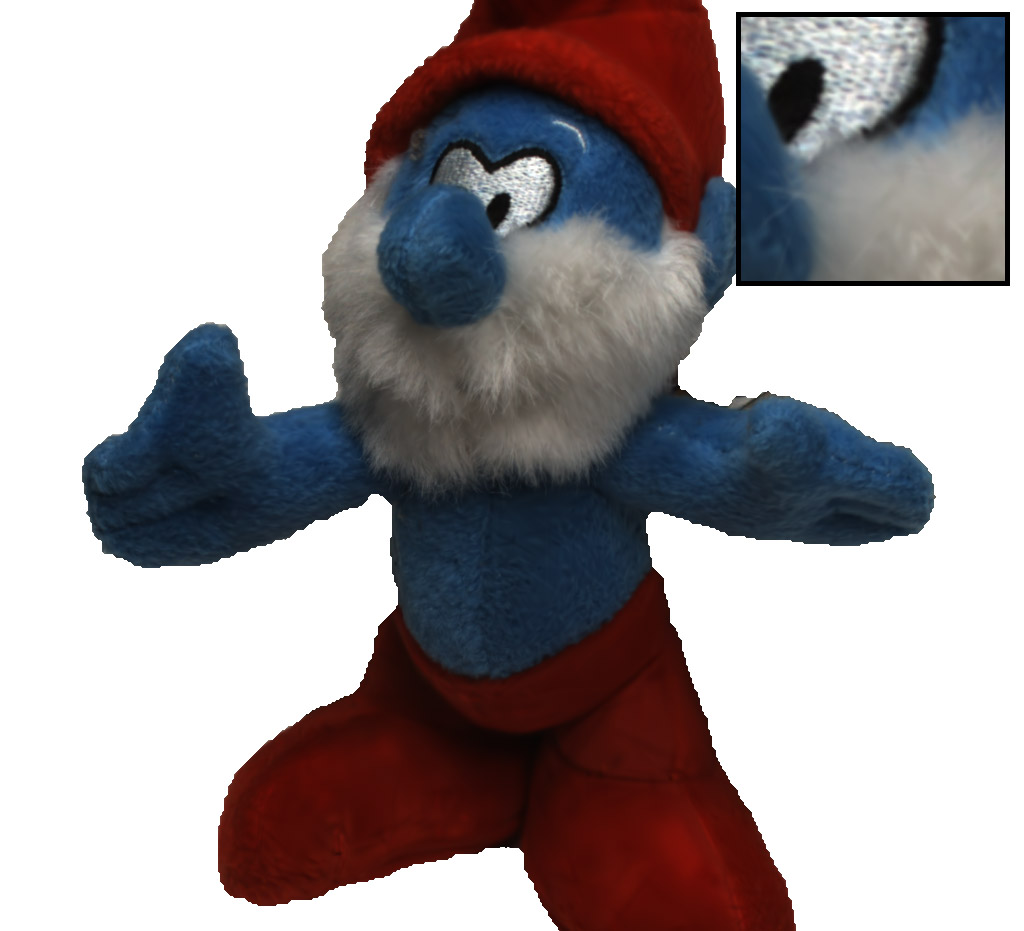} & 
        \includegraphics[width=0.27\textwidth]{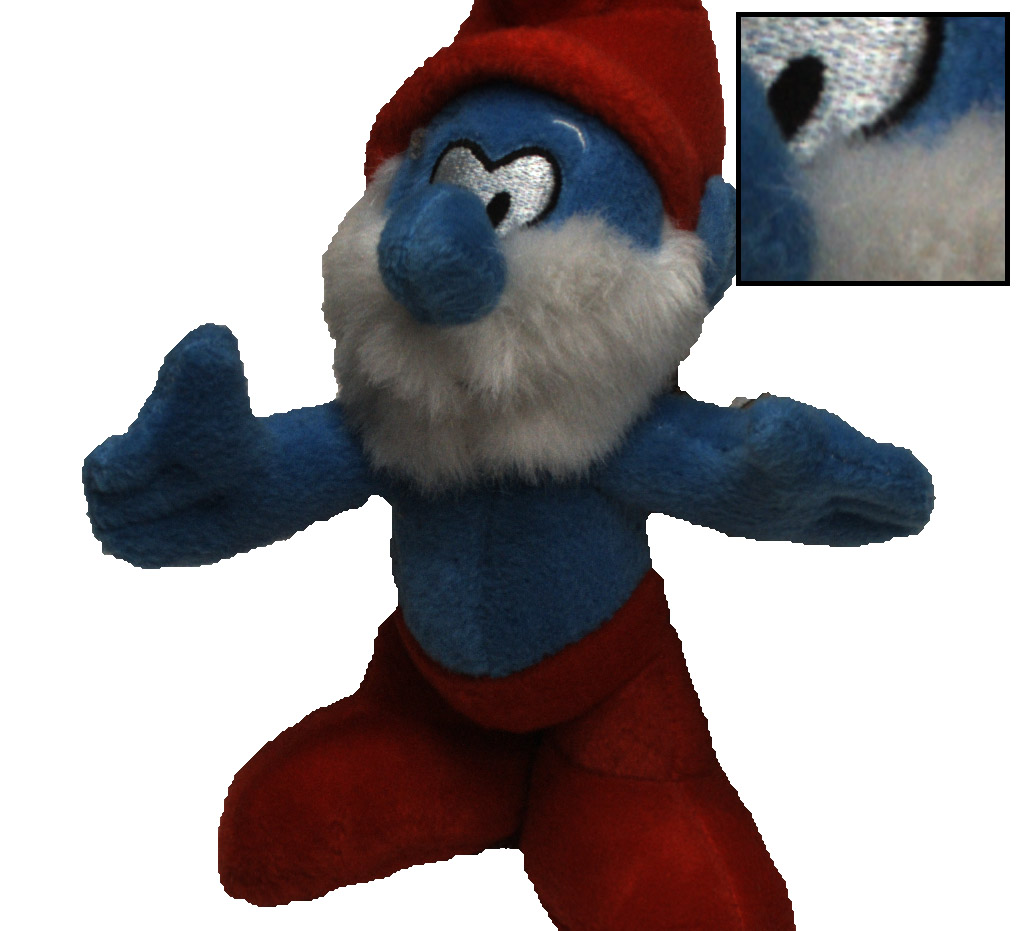} &
        \includegraphics[width=0.27\textwidth]{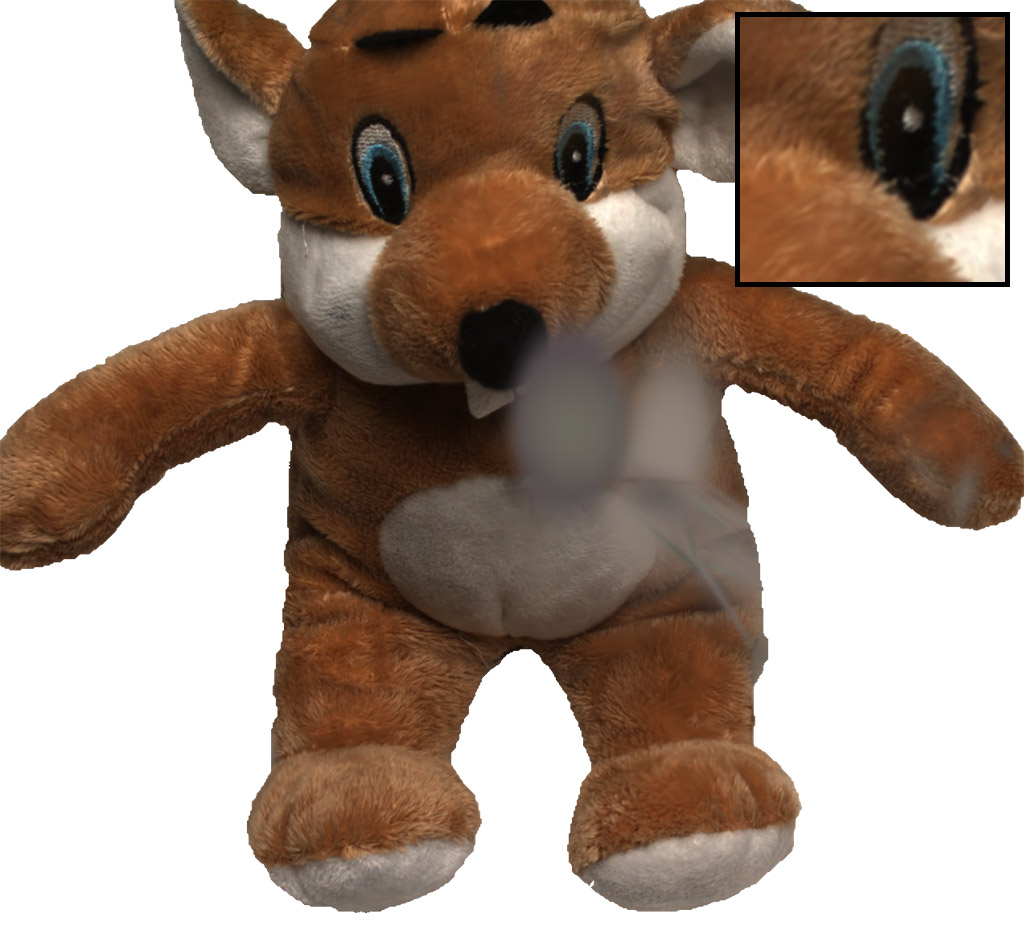} & 
        \includegraphics[width=0.27\textwidth]{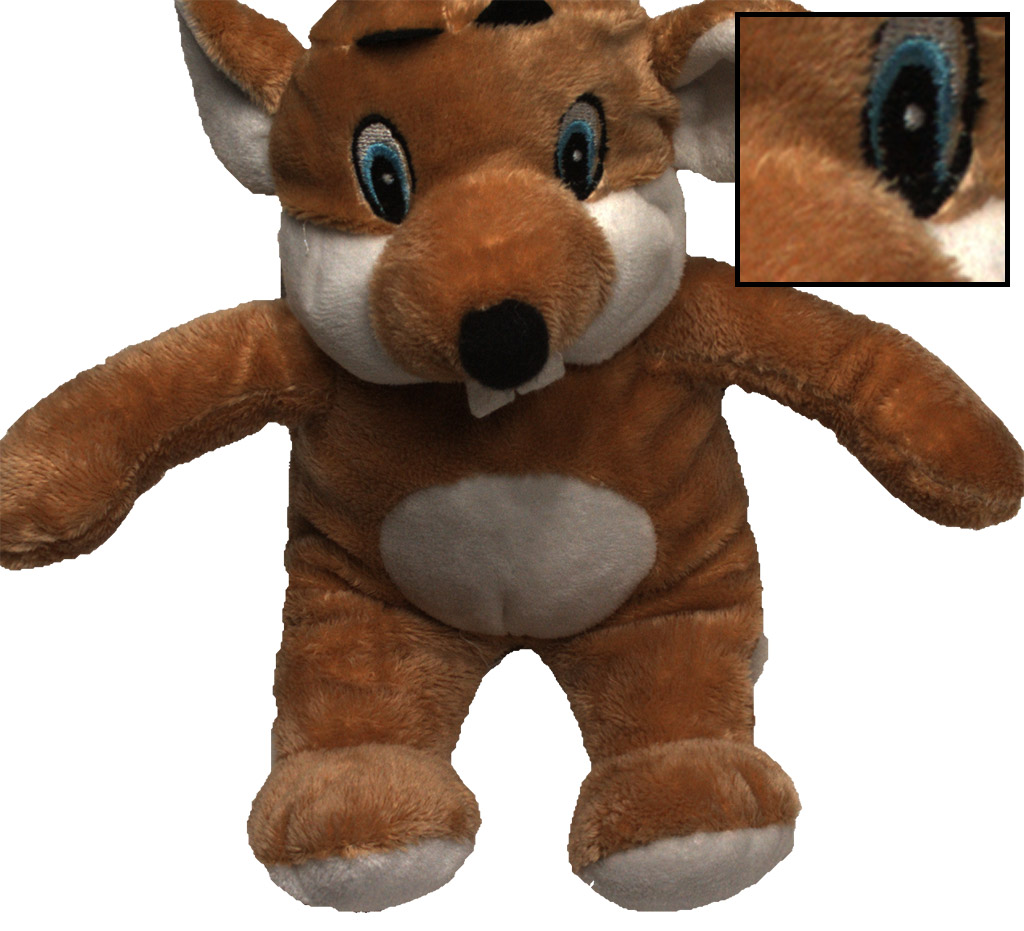} \\
        3DGS & Ground truth & 3DGS & Ground truth 
        
    \end{tabular}
    }
    \caption{
    Qualitative comparison of our 7-Mesh model with PermutoSDF~\cite{Rosu2023CVPR} and 3DGS~\cite{Kerbl2023SIGGRAPH}.
    Scenes from the DTU dataset \cite{jensen2014large}.
    }
    \label{fig:dtu_qualitative}
\end{figure*}

\begin{table*}
\centering
\caption{
Per-scene results for baselines. Refer to \cref{tab:main_results} for averaged results.
Metrics not provided are denoted with “---”.
PermutoSDF trained until densities are fully peaked ($\phi_{\beta_3})$. 
The \textit{hairy monkey} scene is from \citet{Sharma2024ECCV}.
}
\label{tab:scenes_ours_baselines}
\resizebox{\textwidth}{!}{
\begin{tabular}{l|l|ccc|ccc|ccc}
\toprule
\multicolumn{1}{c}{} & \multicolumn{1}{c|}{} & \multicolumn{3}{c|}{PermutoSDF~\cite{Rosu2023CVPR}} & \multicolumn{3}{c}{3DGS~\cite{Kerbl2023SIGGRAPH}} & \multicolumn{3}{c}{MobileNeRF~\cite{chen2023mobilenerf}} \\
\cmidrule(rl){3-11}
\textit{Dataset} & \textit{Scene} & PSNR $\uparrow$ & SSIM $\uparrow$ & LPIPS $\downarrow$
& PSNR $\uparrow$ & SSIM $\uparrow$ & LPIPS $\downarrow$ & PSNR $\uparrow$ & SSIM $\uparrow$ & LPIPS $\downarrow$ \\
\midrule
\multirow{6}{*}{Shelly~\cite{Wang2020SIGGRAPHASIA}} 
& \textit{fernvase} & 28.42 &  0.953 & 0.078 & 34.82 & 0.986 &  0.040 & 29.06 & 0.957 & 0.088 \\ 
& \textit{horse} & 34.68 & 0.993 & 0.040 & 41.45 & 0.997 &  0.038 & 33.31 & 0.988 & 0.065 \\ 
 & \textit{khady} & 26.22 &  0.879 &  0.226 &  30.54 &  0.924 &  0.187 & 26.42 & 0.877 & 0.228 \\ 
& \textit{kitten} & 30.91 & 0.971 & 0.093 & 38.17 & 0.991 & 0.050 & 30.22 & 0.968 & 0.098 \\  
& \textit{pug} & 29.48 & 0.953 &  0.168 & 35.96 &  0.983 &  0.089 & 28.57 & 0.927 & 0.197 \\ 
& \textit{woolly} & 29.39 &  0.949 &  0.167 & 31.71 &  0.969 &  0.130 & 28.20 & 0.919 & 0.221 \\ 
\midrule
\multirow{2}{*}{Custom} 
& \textit{hairy monkey} &  33.67 & 0.977 & 0.194 & 37.67 & 0.990 & 0.142 & 30.25 & 0.949 & 0.200 \\ 
& \textit{plushy} & 32.94 & 0.945 & 0.192 & 37.02 & 0.975 & 0.153 & 31.53 & 0.934 & 0.190 \\  
\midrule
\multirow{2}{*}{DTU~\cite{jensen2014large}} 
& \textit{scan 105} & 34.78 & 0.985 & 0.124 &  35.50 & 0.984 &  0.102 & --- & --- & --- \\ 
& \textit{scan 83} & 37.84 & 0.991 & 0.072 &  40.61 &  0.994 &  0.070 & --- & --- & --- \\ 
\bottomrule
\end{tabular}
} 
\end{table*}



\begin{table*}
\centering
\caption{
Our per-scene results. 
The \textit{hairy monkey} scene is from \citet{Sharma2024ECCV}.
Refer to \cref{tab:main_results} for averaged results.
}
\label{tab:scenes_ours_1}
\begin{tabular}{l|l|ccc|ccc}
\toprule
\multicolumn{1}{c}{} & \multicolumn{1}{c|}{} & \multicolumn{3}{c|}{3-Mesh} & \multicolumn{3}{c}{5-Mesh} \\
\cmidrule(rl){3-8}
\textit{Dataset} & \textit{Scene} & PSNR $\uparrow$ & SSIM $\uparrow$ & LPIPS $\downarrow$
& PSNR $\uparrow$ & SSIM $\uparrow$ & LPIPS $\downarrow$\\
\midrule
\multirow{6}{*}{Shelly~\cite{Wang2020SIGGRAPHASIA}} 
& \textit{fernvase} & 32.41 & 0.985 & 0.066 & 33.63 & 0.988 & 0.064 \\ 
& \textit{horse} & 38.34 & 0.998 & 0.038 & 39.78 & 0.998 & 0.034 \\ 
 & \textit{khady} & 29.78 & 0.938 & 0.193 & 29.88 & 0.941 & 0.194 \\ 
& \textit{kitten} & 35.84 & 0.991 & 0.078 & 36.85 & 0.992 & 0.076 \\ 
& \textit{pug} & 33.72 & 0.983 & 0.138 & 34.25 & 0.985 & 0.132 \\
& \textit{woolly} & 30.26 & 0.973 & 0.175 & 31.12 & 0.978 & 0.162 \\ 
\midrule
\multirow{2}{*}{Custom} 
& \textit{hairy monkey} & 35.59 & 0.984 & 0.178 & 35.90 & 0.985 & 0.179 \\ 
& \textit{plushy} & 34.41 & 0.957 & 0.164 & 34.99 & 0.965 & 0.164 \\ 
\midrule
\multirow{2}{*}{DTU~\cite{jensen2014large}} 
& \textit{scan 105} & 34.77 & 0.980 & 0.120 & 35.40 & 0.982 & 0.106 \\ 
& \textit{scan 83} & 38.05 & 0.990 & 0.064 & 38.34 & 0.990 & 0.063 \\ 
\bottomrule
\end{tabular}
\end{table*}



\begin{table*}
\centering
\caption{
Our per-scene results.
The \textit{hairy monkey} scene is from \citet{Sharma2024ECCV}.
Refer to \cref{tab:main_results} for averaged results.
}
\label{tab:scenes_ours_2}
\begin{tabular}{l|l|ccc|ccc}
\toprule
\multicolumn{1}{c}{} & \multicolumn{1}{c|}{} & \multicolumn{3}{c|}{7-Mesh} & \multicolumn{3}{c}{9-Mesh} \\
\cmidrule(rl){3-8}
\textit{Dataset} & \textit{Scene} & PSNR $\uparrow$ & SSIM $\uparrow$ & LPIPS $\downarrow$
& PSNR $\uparrow$ & SSIM $\uparrow$ & LPIPS $\downarrow$\\
\midrule
\multirow{6}{*}{Shelly~\cite{Wang2020SIGGRAPHASIA}} 
& \textit{fernvase} & 34.55 & 0.990 & 0.062 & 34.64 & 0.991 & 0.062 \\ 
& \textit{horse} & 40.05 & 0.998 & 0.033 & 39.32 & 0.998 & 0.034 \\ 
 & \textit{khady} & 29.97 & 0.942 & 0.194 & 29.96 & 0.943 & 0.195 \\ 
& \textit{kitten} & 37.11 & 0.993 & 0.074 & 37.05 & 0.993 & 0.074 \\ 
& \textit{pug} & 34.25 & 0.985 & 0.132 & 34.24 & 0.985 & 0.133 \\
& \textit{woolly} & 31.04 & 0.978 & 0.158 & 31.05 & 0.978 & 0.160 \\ 
\midrule
\multirow{2}{*}{Custom} 
& \textit{hairy monkey} & 36.09 & 0.987 & 0.177 & 36.14 & 0.987 & 0.175 \\ 
& \textit{plushy} & 35.18 & 0.967 & 0.162 & 35.35 & 0.969 & 0.160 \\ 
\midrule
\multirow{2}{*}{DTU~\cite{jensen2014large}} 
& \textit{scan 105} & 35.50 & 0.982 & 0.106 & 35.54 & 0.983 & 0.105 \\ 
& \textit{scan 83} & 38.04 & 0.991 & 0.062 & 38.81 & 0.991 & 0.062 \\ 
\bottomrule
\end{tabular}
\end{table*}




\begin{table*}
    \centering
    \caption{Results averaged across test scenes.
    In solid scenes, our method outperforms PermutoSDF~\cite{Rosu2023CVPR} (see \cref{fig:blender_qualitative}) but lags behind other baselines. We explain this behavior in \cref{sec:solid-scenes}.
    Methods marked with a $\star$ show results taken from original papers. 
    %
    %
    %
    PermutoSDF trained until densities are fully peaked ($\phi_{\beta_3})$. 
    Metrics not provided by a baseline are denoted with “---”. 
    }
    \label{tab:quantitative_blender}
    \begin{tabular}{l|ccc|ccc}
        \toprule
        & \multicolumn{6}{c}{NeRF-Synthetic~\cite{Mildenhall2020ECCV}}\\
        \cmidrule(rl){2-7}
        & \multicolumn{3}{c|}{\textit{Training}} & \multicolumn{3}{c}{\textit{Test}} \\
        \cmidrule(rl){2-7}
        \textit{Method} & PSNR $\uparrow$ & SSIM $\uparrow$ & LPIPS $\downarrow$ & PSNR $\uparrow$ & SSIM $\uparrow$ & LPIPS $\downarrow$ \\ 
        \midrule
        3DGS~\cite{Kerbl2023SIGGRAPH} & 36.76 & 0.991 &  0.030 & 33.23 & 0.981 &  0.037 \\ 
        Instant-NGP~\cite{Mueller2022SIGGRAPH} $^\star$ &  --- &  --- &  --- &  33.18 &  --- &  --- \\
        PermutoSDF~\cite{Rosu2023CVPR} &  29.31 &  0.975 &  0.057 &  28.05 & 0.966 &  0.065 \\ 
        AdaptiveShells~\cite{Wang2020SIGGRAPHASIA} $^\star$  &  --- &  --- &  --- &  31.84 &  0.957 &  0.056  \\
        QuadFields~\cite{Sharma2024ECCV} $^\star$  &  --- &  --- &  --- &  31.00 &  0.952 &  0.069 \\
        MobileNeRF~\cite{chen2023mobilenerf} $^\star$ &  --- &  --- &  --- &  30.90 &  0.947 &  0.060  \\
        \midrule
        3-Mesh &  32.40 &  0.983 &  0.060 &  28.50 &  0.958 &  0.083 \\ 
        5-Mesh &  33.23	&  0.986 &  0.055 &  28.77 &  0.959	&  0.081 \\
        7-Mesh &  33.31	&  0.986 &  0.056 &  28.88 &  0.960 &  0.081 \\
        9-Mesh &  33.19 &  0.986 &  0.057 &  28.79 &  0.960	&  0.082 \\ 
        \bottomrule
        \end{tabular}
\end{table*}

\end{document}